\pdfoutput=1

\documentclass[11pt]{article}

\usepackage{acl}

\usepackage{times}
\usepackage{latexsym}
\usepackage{xcolor} 
\usepackage{amssymb}
\usepackage{booktabs}
\usepackage{array}
\usepackage{multirow}
\usepackage{graphicx}
\usepackage{subcaption}
\usepackage{algorithm}
\usepackage{algorithmic}
\usepackage{float}
\usepackage{amsmath}
\usepackage{enumitem}   
\usepackage{tikz}
\usepackage{pgfplots}
\usepackage{cuted}   
\usepackage{capt-of} 
\pgfplotsset{compat=1.18}
\usepackage{diagbox}
\usepackage{amssymb}

\usepackage[T1]{fontenc}

\usepackage[utf8]{inputenc}

\usepackage{microtype}

\usepackage{inconsolata}

\usepackage{graphicx}
\title{Revealing Side Effects of Using Style Features on Conversational Agents}

\author{First Author \\
  Affiliation / Address line 1 \\
  Affiliation / Address line 2 \\
  Affiliation / Address line 3 \\
  \texttt{email@domain} \\\And
  Second Author \\
  Affiliation / Address line 1 \\
  Affiliation / Address line 2 \\
  Affiliation / Address line 3 \\
  \texttt{email@domain} \\}

\newcommand{\promptText}[1]{\textcolor{orange}{\textit{#1}}}

\newcommand{\gptFiveMini}{\texttt{GPT-5-mini}}

\newcommand{\llamaThreeOne}{\texttt{Llama-3.1-8B-Instruct}}
\newcommand{\qwenNineB}{\texttt{Qwen3.5-9B}}
\newcommand{\qwenTwentySevenB}{\texttt{Qwen3.5-27B}}
\newcommand{\gemmaThirtyOneB}{\texttt{Gemma-4-31B}}
\newcommand{\gemmaEfourB}{\texttt{Gemma-4-E4B-it}}
\newcommand{\gptOssTwentyB}{\texttt{GPT-OSS-20B}}
\newcommand{\gptOssOneTwentyB}{\texttt{GPT-OSS-120B}}
\newcommand{\geminiFlashLite}{\texttt{Gemini-3.1-Flash-Lite}}
\newcommand{\llamaThreeOneAbbr}{\texttt{Llama3.1-8B}}
\newcommand{\qwenNineBAbbr}{\texttt{Qwen3.5-9B}}
\newcommand{\qwenTwentySevenBAbbr}{\texttt{Qwen3.5-27B}}
\newcommand{\gemmaThirtyOneBAbbr}{\texttt{Gemma4-31B}}
\newcommand{\gemmaEfourBAbbr}{\texttt{Gemma4-E4B}}
\newcommand{\gptOssTwentyBAbbr}{\texttt{GPT-OSS-20B}}

\newcommand{\gptFiveMiniAbbr}{\texttt{GPT-5-mini}}
\newcommand{\geminiFlashLiteAbbr}{\texttt{Gemini3.1-FL}}
\newcommand{\mainFeature}{Main Feature}
\newcommand{\sideFeature}{Side Feature}
\newcommand{\evaluationFeature}{Evaluation Feature}
\newcommand{\code}[1]{\texttt{#1}}

\newcolumntype{C}[1]{>{\centering\arraybackslash}p{#1}} 
\newcolumntype{L}[1]{>{\raggedright\arraybackslash}p{#1}}

\author{
 \textbf{Young-Min Cho \textsuperscript{1,2}\thanks{Equal contribution. This work was done at the University of Pennsylvania.}},
 \textbf{Yuan Yuan\textsuperscript{1,3}\footnotemark[1]},
 \textbf{Sharath Chandra Guntuku \textsuperscript{1}},
 \textbf{Lyle Ungar\textsuperscript{1}},
\\
 \textsuperscript{1}University of Pennsylvania,
 \textsuperscript{2}AWS,
 \textsuperscript{3}University of California, Los Angeles
\\
 \small{
   \href{mailto:ymcho@amazon.com}{ymcho@amazon.com}, 
   \href{mailto:yyuan20@g.ucla.edu}{yyuan20@g.ucla.edu}
 }
}

\begin{document}
\maketitle
\begin{abstract}
Style instructions such as \textit{helpful}, \textit{friendly}, and \textit{professional} are widely used to steer LLM-based conversational agents, but targeting one style can alter unprompted dimensions not specified in the instruction. We call these changes \textbf{stylistic side effects}. To ground our study in practice, we survey 125 conversational-agent papers from the ACL Anthology and identify 13 frequently used style features. We then evaluate eight models using controlled synthetic dialogues in task-oriented and open-domain settings, measuring each prompt’s cross-style effects against neutral responses. Across models and domains, we identify systematic cross-style side effects, showing that commonly used style prompts do not control stylistic dimensions independently. We also evaluate four mitigation strategies: Prompting, Steering, DPO, and SFT. Although these methods often move affected features in the desired direction, they commonly weaken the originally targeted style. Response-level analysis suggests that this trade-off reflects bundled style realizations: lexical and content choices expressing the target style also shift other judged dimensions. These findings challenge the assumption that natural-language style prompts provide faithful, isolated control and motivate multi-objective methods for reliable stylistic steering. \footnote{The survey annotations, experimental results, and prompt templates are available at \url{https://github.com/JeffreyCh0/style-side-effects}.}
\end{abstract}

\section{Introduction} \label{sec:introduction}

The use of natural language prompts to steer the persona and output style of LLMs is now a ubiquitous practice in conversational agent design. Developers and researchers routinely employ system instructions, such as "be friendly," "be professional," or "be helpful", to tailor model behaviors for specific applications ranging from mental health support to customer service \cite{feng-etal-2025-emocharacter, zhao-etal-2025-diversity, rachidi-etal-2025-design}. These prompt-based controls often implicitly assume that stylistic dimensions can be manipulated independently, without accounting for potential interactions among correlated traits. In practice, prompting for a particular style may activate a constellation of associated traits, shaped by training data, social conventions, and latent representations within the model.

In this work, we interrogate this entanglement and identify a pervasive phenomenon we term \textit{stylistic side effects}: statistically significant behavioral shifts in unprompted styles whose directions relative to the neutral baseline conflict with a semantic-similarity prior. While prior work examines prompt side effects and persona-induced behavioral changes \cite{Luz_de_Araujo_2025, gupta2024biasrunsdeepimplicit, kang-hovy-2021-style}, systematic evidence on how commonly used stylistic controls affect one another remains limited.

We address this gap by presenting a systematic study of cross-feature stylistic side effects in LLMs. To ground our analysis in real-world usage, we first conduct a comprehensive survey of 125 conversational agent papers from the ACL Anthology (2023–2025), identifying 13 distinct, frequently used style features such as \textit{helpful}, \textit{empathetic}, and \textit{friendly}. We then implement a controlled evaluation framework, generating synthetic dialogues across task-oriented and open-domain settings using eight models spanning local open-weight and API-served systems. By employing an LLM-as-a-Judge pairwise comparison protocol with win rates~\citep{zheng2023judging}, we quantify how prompting for a \mainFeature{} impacts the expression of other, unprompted \sideFeature{}s.


Our results reveal that style features are deeply entangled. We demonstrate consistent and structured side effect patterns; for example, prompting for \textit{Friendly} significantly reduces perceived \textit{Professional} in daily conversations across six of the eight models.
Furthermore, we evaluate four mitigation strategies: Prompting, Steering, Direct Preference Optimization (DPO), and Supervised-Finetuning (SFT). In our tested settings, we find that attempting to restore suppressed traits often degrades the primary intended style, highlighting the need for more principled, multi-objective approaches to reliable stylistic steering.
    
\begin{itemize}
    \item \textbf{We conduct a systematic survey} of style-feature usage in 125 conversational-agent papers from the ACL Anthology (2023--2025), deriving a 13-feature inventory of frequently used style controls for controlled evaluation.
    
    \item \textbf{We characterize cross-feature stylistic effects} by estimating how prompting a Main Feature changes the judged expression of other, unprompted Side Features across eight models and two dialogue domains, showing structured output-level dependencies among style controls.
    
    \item \textbf{We evaluate four mitigation strategies}: Prompting, Steering, DPO, and SFT, and identify a recurring trade-off between correcting selected Side Feature shifts and preserving the prompted Main Feature.
\end{itemize}

\section{Related Work}
\label{sec:related_work}
Recent research on persona-oriented and personalized LLMs highlights the importance of stylistic control in interactive settings such as role-playing and long-term user adaptation \citep{tseng2024talespersonallmssurvey, chevi2025how}.\footnote{Since style is closely related to broader notions such as persona, personalization, and role-playing, we review these areas insofar as they require reliable control over stylistic behavior.} Style also affects safety, robustness, and evaluation reliability: superficial style alignment can weaken safety guardrails \citep{xiao2025when}, while stylistic biases can lead LLM judges to favor pleasing presentation \citep{feuer2025style}. Complementary benchmarks such as PersonalLLM \citep{zollo2025personalllm}, PERSONAMEM \citep{jiang2025know, jiang2025personamem}, and Crab \citep{he2025crab} evaluate whether models maintain coherent behavior across users, roles, and interaction histories. Model brittleness under stylistic perturbations \citep{truong2025persona} further motivates style-aware optimization that preserves effectiveness across stylistic contexts \citep{pu2024style}.

Prior work also shows that style and persona attributes are multidimensional and interacting rather than independently controllable. \citet{kang-hovy-2021-style} describe style as a set of covarying factors, while \citet{labroo2026funny} show that combining even seemingly separable concepts can weaken individual control; related studies likewise report conflicts and unintended shifts in multi-attribute generation and persona steering \citep{cao2024tara, de2024dynamic, chen2025persona, elgaar2026linggen}. Prior works suggest that adding or strengthening one attribute can alter dimensions that are not explicit optimization targets. We extend this line of research by systematically identifying and characterizing unprompted cross-style shifts induced by a single style instruction, and by examining the trade-off between correcting the affected features and preserving the intended style.

LLM stylistic steering commonly relies on natural-language instructions such as “be empathetic” \citep{feng-etal-2025-emocharacter, zhao-etal-2025-diversity, rachidi-etal-2025-design}. Although flexible and parameter-free, prompting can be superficial or counterproductive: expert personas may degrade objective-task performance \citep{zheng2024ahelpfulassistantreally}, and persona or style prompts may activate latent stereotypes rather than precise behavioral constraints \citep{lutz2025promptmakespersonasystematic, malik-etal-2024-empirical}. Activation-steering methods such as CAA offer alternatives without full-model fine-tuning \citep{panickssery2024steeringllama2contrastive, zhang-etal-2025-personalized}; subsequent work aligns personality traits through preference-based steering \citep{bo2025steerable} and proposes more disentangled controls such as SteerX, StyleVector, and PersonaVector \citep{zhao2025steerx, zhang-etal-2025-personalized, chen2025persona}. SFT and DPO have also been applied to character-specific role-playing \citep{shao2023character}, Big Five personality induction \citep{li2025big5}, persona-faithful role-playing \citep{peng2024quantifying}, and role-style instruction following \citep{lu2025rolemrc}. Together, these methods advance controllable adaptation but leave open how to manage interactions among style, task content, user preferences, safety behavior, and latent stereotypes.

\section{Style Feature Extraction - A Survey}
\label{sec:style_feature_extract}
To ground our study in contemporary practice, we perform a systematic survey of conversational-agent papers published in the \textit{ACL Anthology} between 2023 and 2025, following the PRISMA framework~\citep{tugwell2021prisma} for literature identification, screening, and selection.
We extract commonly used style features from the selected papers. Further details of the survey procedure and feature inventory are provided in Appendix~\ref{appendix:style_feature_extraction_survey_details}.
    
We searched the ACL Anthology for papers published from January 2023 through June 2025 whose titles or abstracts contained one or more of the following terms: `conversational agent', `dialogue system', `dialog system', and `chatbot'. We screened 586 papers at full text and extracted style features from the agent prompts in 125 papers. We normalized the extracted features to adjectives and retained features that appeared in at least five distinct papers, yielding 16 candidates.

Because we use length-controlled generation to minimize the influence of length on LLM judge, we excluded \textit{concise}, \textit{detailed}, and \textit{short}.
As the result, we extracted 13 distinct style features: \textit{efficient}, \textit{impartial}, \textit{expert}, \textit{helpful}, \textit{outgoing}, \textit{friendly}, \textit{curious}, \textit{polite}, \textit{engaging}, \textit{empathetic}, \textit{professional}, \textit{informative}, and \textit{caring}, that typify how recent papers employ prompt‑based message control.

\section{Data Generation and Annotation}
\label{sec:generation_pipeline}
In this section, we introduce a controlled generation and evaluation pipeline for estimating how explicit style feature prompts shift an agent's judged stylistic behavior. Let $d\in\mathcal D$ index a dialogue domain, $g\in\mathcal G$ a generation model, $q_i$ an initiating message, and $a,b\in\mathcal F$ style features. We call $a$ the \textbf{\mainFeature{}}, because it is explicitly requested in the generation prompt, and $b$ the \textbf{\evaluationFeature{}}, because it is the axis on which the response pair is judged. The diagonal case $b=a$ measures the intended feature effect. We reserve \textbf{\sideFeature{}} for an unprompted evaluation feature, that is, the off-diagonal case $b\ne a$. The goal of the pipeline is to measure both the intended effect and shifts along other stylistic dimensions.

To cover diverse, real-world domains for message generation, we use two dialogue datasets. LMSYS-Chat-1M (Task, \citealt{zheng2023lmsyschat1m}) represents task-oriented inquiries, while DailyDialog (Daily, \citealt{li-etal-2017-dailydialog}) represents open-domain, daily-life interactions. For each domain, we select 10 topics and sample 90 initial messages per topic; topic construction and sampling are detailed in Appendix~\ref{sec:selection_of_topics}, and the topics and example messages appear in Appendix Table~\ref{tab:domain_topic_examples}. Of these 90 messages, 10 are reserved for side effect identification and 80 for intervention experiments, with no overlap between the two subsets. Each conversation is initiated by a single human message.

We generate responses using LLMs that span a range of model families, sizes, and access modes. Six models are local open-weight models: \llamaThreeOne{} (\texttt{Llama3.1-8B}, \citealp{grattafiori2024llama3herdmodels}), \qwenNineB{} and \qwenTwentySevenB{} \citep{qwen2026qwen3_5}, \gemmaThirtyOneB{} (\texttt{Gemma4-31B}) and \gemmaEfourB{} (\texttt{Gemma4-E4B}, \citealp{googledeepmind2026gemma4}), and \gptOssTwentyB{} \cite{openai2025gptoss120bgptoss20bmodel}. Two additional models are accessed through hosted APIs: \gptFiveMini{} \cite{openai2025gpt5} and \geminiFlashLite{} (\texttt{Gemini3.1-FL}, \citealp{googledeepmind2026gemini}). This model set allows us to evaluate whether style-induced shifts are consistent across both local and API-served systems.

For response generation, we define two prompting conditions. In the \textit{Styled} condition, a style feature is applied by querying the agent with the system prompt: \promptText{"You are having a conversation about \{topic\}. Please be \{style\_feature\} in your response."} In the control condition, denoted \textit{Neutral}, we use the same topic framing but remove the stylistic instruction: \promptText{"You are having a conversation about \{topic\}."} All generations use relatively high temperatures to provide sufficient variation for estimating stylistic effects.\footnote{Local models are sampled with \texttt{temperature}\(=0.7\), while API-served models, \gptFiveMini{} and \geminiFlashLite{}, use \texttt{temperature}\(=1.0\). For models that expose a reasoning-level knob, we set it to its lowest available level.} To obtain a clean estimate and minimize intermediate confounding, we only record the agent's immediate response to the initiating message for data collection. 

Because response length can confound pairwise style judgments, we match responses in expected length on a per-message basis. For each initial message $q_i$ and model $g$, we first generate five \textit{Neutral} responses without a length cap. We use the rounded mean word count of these responses as the per-message, per-model target length $L_{q_i,g}$. We then append the instruction \promptText{"Please answer in around \{$L_{q_i,g}$\} words."} to every subsequent system prompt, including both the \textit{Neutral} baseline prompt and all \textit{Styled} prompts. Appendix Figure~\ref{fig:length_distribution} shows that the length-controlled \textit{Neutral} and \textit{Styled} responses occupy similar length regions around the per-message budgets, suggesting that the procedure substantially reduces systematic response-length differences between conditions.

We evaluate stylistic shifts using a pairwise comparison protocol. Each comparison pair consists of a \textit{Styled} response generated with a given \mainFeature{} and its corresponding \textit{Neutral} response generated without that \mainFeature{}. The pair is evaluated with respect to a target \evaluationFeature{}, asking which response exhibits that feature more strongly. All primary pairwise judgments throughout the paper are produced by a single fixed judge, \gptOssOneTwentyB{} \cite{openai2025gptoss120bgptoss20bmodel}. During evaluation, response order is randomized, and the judge must identify which response more strongly expresses the \evaluationFeature{}; the exact prompt and a cross-judge sensitivity analysis are provided in Appendix~\ref{appendix:generation_evaluation_protocols}. We use \textbf{win rate} as the primary metric: values above chance indicate that prompting a \mainFeature{} increases the expression of a given \evaluationFeature{}, while values below chance indicate a decrease.

\begin{figure}[t!]
    \centering
    \includegraphics[width=\linewidth]{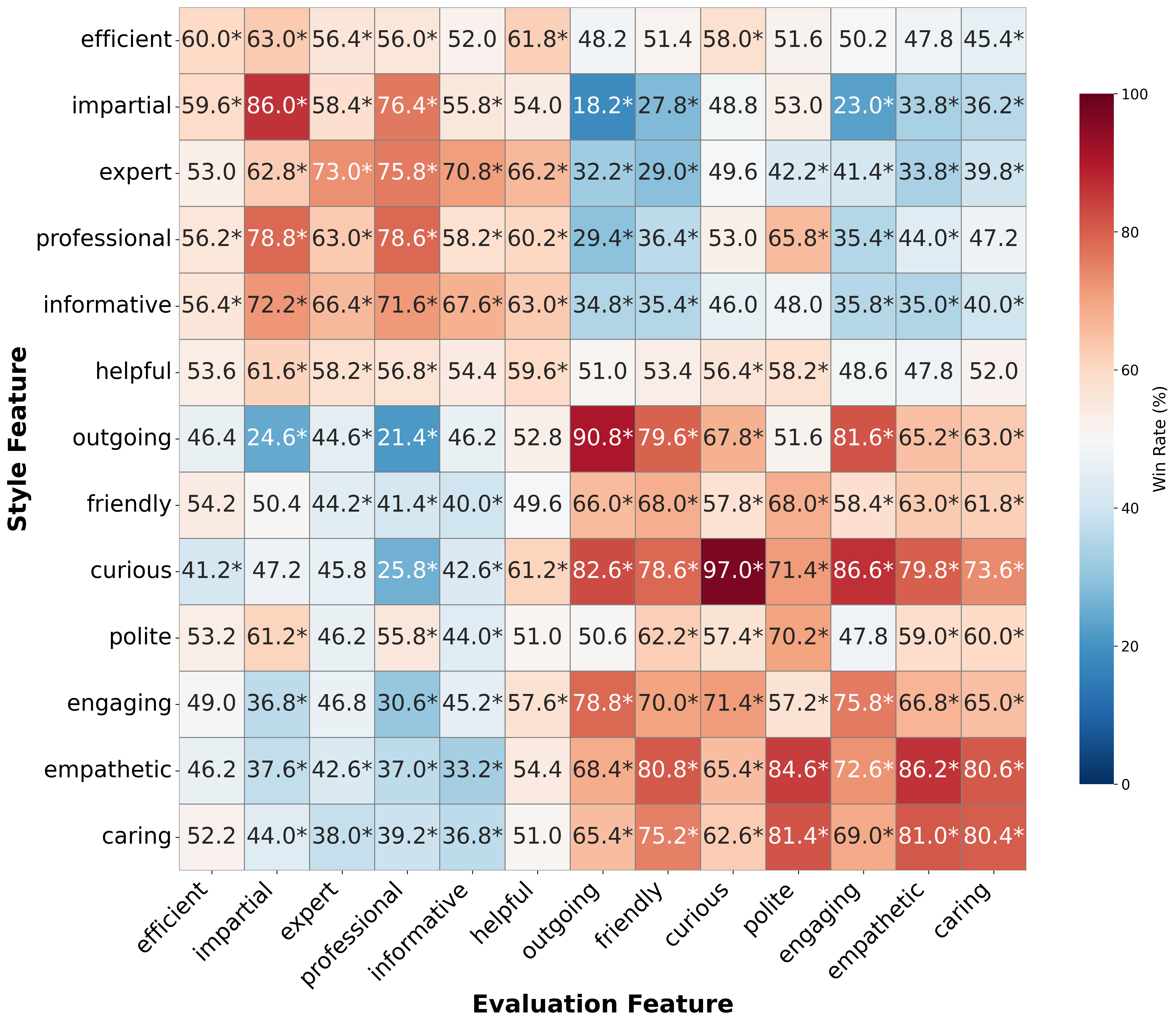}
    \caption{Style feature effect matrix for \gptFiveMini{} (daily domain). Rows are prompted \mainFeature{}s, and columns are judged \evaluationFeature{}s. Each cell reports \(100\cdot W(a,b)\%\) for the styled response over the neutral baseline; * marks \(p_W(a,b) \le 0.05\) (binomial).}
    \label{fig:correlation_matrix}
\end{figure}

\section{Style Feature Effect Analysis}
\label{correlation results}

To examine how applying a \mainFeature{} shifts model behavior along the \evaluationFeature{} axis, we computed win rates for all pairwise combinations of 13 \mainFeature{}s and 13 \evaluationFeature{}s. For each message--\mainFeature{} pair, responses were sampled five times. We then aggregated these win rates to form a \textit{style feature
effect matrix}, where rows correspond to prompted \mainFeature{}s 
and columns correspond to evaluated \evaluationFeature{}s.

This matrix measures the extent to which a particular style feature, such as \textit{Friendly}, biases an agent toward specific stylistic dimensions, such as \textit{Expert} or \textit{Helpful}. In this way, the full win-rate matrix captures how explicitly imposing one style feature influences model behavior across a broader space of stylistic attributes. The color scale represents the average win rate: values above 50\%, shown in red, indicate that the styled response exhibits a stronger tendency toward the corresponding \evaluationFeature{}, whereas values below 50\%, shown in blue, indicate a negative effect.

We present the \textit{style feature effect matrix} for \gptFiveMini{} on the daily domain in Figure~\ref{fig:correlation_matrix} as an example; Appendix Figure~\ref{fig:combined_heatmap} reports the full matrices for all eight models in both domains. Across these matrices, the red diagonal shows that style prompting steers the model toward the intended feature effects. Beyond this primary effect, we observe that features naturally cluster by feature type: utility-oriented features (e.g., \textit{Expert}, \textit{Professional}, \textit{Informative}, \textit{Helpful}) consistently show high win rates against one another, and the same holds for interpersonal features (e.g., \textit{Friendly}, \textit{Polite}, \textit{Engaging}, \textit{Empathetic}). Crucially, the matrices also reveal significant \textit{side effects}, defined algorithmically in Section~\ref{sec:identifying_side_effects}, across the two clusters: prompting the model to be \textit{Friendly}, for instance, makes it noticeably less \textit{Professional}.

\begin{table}[t!]
    \centering
    \small
    \resizebox{\columnwidth}{!}{
        \begin{tabular}{lllll}
            \toprule
            \textbf{Domain} & \textbf{Model} & \textbf{Main Feature} & \textbf{Effect} & \textbf{Side Feature} \\
            \midrule
            \multirow{8}{*}{Daily}
              & \texttt{Gemini3.1-FL}   & Friendly   & Less & Professional \\
              & \texttt{Gemma4-31B}     & Friendly   & Less & Professional \\
              & \texttt{Gemma4-E4B}     & Friendly   & Less & Professional \\
              & \texttt{GPT-5-mini}     & Empathetic & More & Outgoing \\
              & \texttt{GPT-OSS-20B}    & Friendly   & Less & Professional \\
              & \texttt{Llama3.1-8B}    & Curious    & Less & Informative \\
              & \texttt{Qwen3.5-9B}     & Friendly   & Less & Professional \\
              & \texttt{Qwen3.5-27B}    & Friendly   & Less & Professional \\
            \midrule
            \multirow{8}{*}{Task}
              & \texttt{Gemini3.1-FL}   & Impartial    & More & Professional \\
              & \texttt{Gemma4-31B}     & Impartial    & More & Professional \\
              & \texttt{Gemma4-E4B}     & Expert       & Less & Efficient \\
              & \texttt{GPT-5-mini}     & Professional & More & Empathetic \\
              & \texttt{GPT-OSS-20B}    & Professional & Less & Friendly \\
              & \texttt{Llama3.1-8B}    & Professional & Less & Friendly \\
              & \texttt{Qwen3.5-9B}     & Professional & Less & Friendly \\
              & \texttt{Qwen3.5-27B}    & Professional & Less & Friendly \\
            \bottomrule
        \end{tabular}
    }
    \caption{Highest-SES side-effect pair selected for each model--domain setting. ``Less'' means the Main Feature suppresses the Side Feature, and vice versa.}
    \label{tab:selected_side_effects}
\end{table}

\section{Side Effect Identification}
\label{sec:identifying_side_effects}

From the treatment effect analysis, we observe that applying a style feature can induce shifts in other, unprompted stylistic dimensions. We define a \textbf{side effect} as a statistically significant change in the judged expression of an unprompted Side Feature that occurs when prompting for a Main Feature, relative to the \textit{Neutral} condition, and whose direction conflicts with the semantic-similarity relationship between the two feature names. The Side Effect Score (SES) ranks qualifying shifts using the empirical--semantic disagreement and the magnitudes of both standardised signals.

\paragraph{Side Effect Score (SES).}
To rank candidate (\mainFeature{}, \sideFeature{}) pairs, we {mechanically} define a Side Effect Score that captures \emph{how strongly} prompting a Main Feature $a$ moves the response on a Side Feature axis $b$, relative to what the prior linguistic similarity between $a$ and $b$ would predict. Let $W(a,b) \in [0,1]$ be the pairwise win rate of the styled response over the neutral baseline on the $b$-axis (judged by \gptOssOneTwentyB{}), and $C(a,b) \in [-1, 1]$ the cosine similarity of the OpenAI \texttt{text-embedding-3-large} \cite{openai_text_embedding_3_large} embeddings of the two feature names. We standardise both quantities using the Median Absolute Deviation: $N_{\textsc{wr}}$ centred at $0.5$, $Z_{\textsc{cs}}$ centred at the off-diagonal mean, and define
\begin{equation*}
\begin{aligned}
\mathrm{SES}(a,b) \;=\;& \bigl| N_{\textsc{wr}}(a,b) - Z_{\textsc{cs}}(a,b) \bigr| \\
\;\cdot\;& \min\bigl(|N_{\textsc{wr}}(a,b)|,\,|Z_{\textsc{cs}}(a,b)|\bigr) \\
\;\cdot\;& \mathbf{1}[\,N_{\textsc{wr}}(a,b)\,Z_{\textsc{cs}}(a,b) < 0\,] \\
\;\cdot\;& \mathbf{1}[\,p_W(a,b) \le 0.05\,].
\end{aligned}
\end{equation*}

The term $\bigl|N_{\textsc{wr}}(a,b) - Z_{\textsc{cs}}(a,b)\bigr|$ measures empirical--semantic disagreement, while $\min\bigl(|N_{\textsc{wr}}(a,b)|, |Z_{\textsc{cs}}(a,b)|\bigr)$ acts as a magnitude floor that downweights cases where either signal is near zero. The score is non-zero only when (i)~the empirical win-rate effect moves \emph{opposite} to what semantic similarity predicts, and (ii)~the win rate is significantly different from $0.5$ ($p_W(a,b) \le 0.05$, two-sided binomial). The sign of $N_{\textsc{wr}}(a,b)$ supplies the intervention polarity: positive marks a gain on $b$ (\textit{suppress\_positive}), while negative marks a loss (\textit{enhance\_negative}). Exact standardization, gating, domain-specific selection rules, and multiple-comparison sensitivity are provided in Appendix~\ref{appendix:side_effect_identification_details}. Note that our contribution here is to mechanically characterize prompt-induced cross-feature shifts, not the subjective "expectedness" of features' correlation.

{ For example, consider the \textit{Friendly}--\textit{Professional} pair for \llamaThreeOneAbbr{} in the Daily domain. Across 500 comparisons, \textit{Friendly}-prompted responses achieve a \textit{Professional} win rate of $37.6\%$ against the \textit{Neutral} baseline. The standardized terms are $N_{\textsc{wr}}=-0.789$ and $Z_{\textsc{cs}}=1.467$: the empirical effect is negative despite the above-average semantic similarity between the feature names. Their signs therefore disagree, and the significance gate also passes ($p_W=3.24\times10^{-8}$, two-sided binomial). The disagreement term is $|-0.789-1.467|=2.256$, and the magnitude floor is $\min(0.789,1.467)=0.789$, yielding $\mathrm{SES}=2.256\times0.789\approx1.780$. The negative win-rate term identifies a loss of \textit{Professional}, corresponding to the \textit{enhance\_negative} intervention polarity.}

Applying SES to these full matrices, we restrict \mainFeature{} to the corresponding feature category in each domain: utility-oriented features for the task domain and interpersonal features for the daily domain. For each model and domain, we rank candidate pairs by SES and choose the top-ranked pair as the intervention target. The resulting 16 selected pairs, corresponding to \(8\) models across \(2\) domains, are reported in Table~\ref{tab:selected_side_effects}.

{ A systematic quantification of feature independence and intersection shows strong effects on Side Features. Using mean absolute deviation, Holm-corrected significance tests, and linear regression in Appendix~\ref{appendix:feature_independence_quantification}, we analyzed 156 off-diagonal feature pairs in each of the 16 model--domain matrices. The mean absolute win-rate deviation from $50\%$ was 16.6 percentage points; on average, $59.1\%$ of pairs deviated by at least 10 points, and $55.8\%$ met this threshold with Holm-corrected significance. Semantic similarity explained an average of $12.9\%$ of the win-rate variance. These results show that substantial Side Features' effects extend beyond the selected high-SES pairs, while much of the variation remains unexplained by the similarity-based regression.}

\section{Side Effect Mitigation}
\label{sec:mitigation}

For each SES-selected pair, the goal of Side Effect Mitigation (Mitigation) is to counteract the shift in the \sideFeature{} while preserving the strength of the explicitly prompted \mainFeature{}. { Past papers have shown such applications are needed: for example, \citet{lechner-etal-2023-challenges} prompt an AI doctor to be both \textit{Friendly} and \textit{Professional}, one of the \mainFeature{}-\sideFeature{} pairs found in Table \ref{tab:selected_side_effects}.} To address the trade-off, we evaluate four mitigation methods: Prompting, Steering, DPO, and SFT. We analyze whether they move the \sideFeature{} in the targeted intervention direction without sacrificing the prompted stylistic objective. Steering pseudocode and per-model training configurations are provided in Appendix~\ref{appendix:side_effect_mitigation_details}.

 We prepare the experiment dataset using the intervention set described in Section~\ref{sec:generation_pipeline}. Experiments are domain-specific: each trained model uses data from only one domain. For each domain, we collect $800$ initial messages, consisting of $80$ messages per topic across $10$ topics. We split these messages with stratified sampling by topic into train/validation/test splits using a $6{:}2{:}2$ ratio, giving $48$/$16$/$16$ messages per topic and $480$/$160$/$160$ messages per domain. For training, each first-message is sampled $10$ times, yielding $4{,}800$ training examples. For validation and testing, each first-message is sampled $5$ times, yielding $800$ validation and $800$ test examples. Prompting does not use the training split, but it performs inference only on the test split. 
 
We use \textbf{win rate} as the primary metric in this experiment. Following a procedure similar to that described in Section~\ref{sec:generation_pipeline}, each generated message is compared against its corresponding \textit{Neutral} response. To further compare different methods, we introduce the following metrics. \textbf{\(\Delta_{\mathrm{Main}}\)} and \textbf{\(\Delta_{\mathrm{Side}}\)} denote the mean percentage-point changes in the \mainFeature{} and \sideFeature{} win rates, respectively, where \textbf{\(\Delta_{\mathrm{Side}}\)} is measured in the desired direction. \textbf{Main Flip} measures the proportion of cases in which mitigation moves the \mainFeature{} win rate across the 50\% threshold, thereby rendering the mitigation ineffective. \textbf{Side Flip} measures the proportion of cases in which the \sideFeature{} win rate ends up on the desired side of 50\% after mitigation, indicating successful mitigation.
 
\subsection{Methods}
\paragraph{Prompting}
As a simple mitigation strategy, we modify the system instruction to acknowledge the \sideFeature{} explicitly, with a template that depends on the polarity of the side effect (Section~\ref{sec:identifying_side_effects}). 
For \textit{suppress\_positive} cases (the Main Feature elevates the Side Feature), we use \promptText{"Please be \{main\} but not \{side\}"}; for \textit{enhance\_negative} cases (the Main Feature suppresses the Side Feature), we use \promptText{"Please be \{main\} but \{side\}."}

\paragraph{Steering}
We extract per-layer steering vectors via Contrastive Activation Addition (CAA, \citealp{panickssery2024steeringllama2contrastive}): for each style feature we contrast activations elicited by paired \textit{Styled}-vs-\textit{Neutral} prompts, average the difference at a fixed token position over the training split, and pick the injection layer that has the strongest style feature effect on the validation split. At inference we condition the model on the prompt \promptText{"Please be \{main\} in your response."} and add the \sideFeature{} vector (or its negative, depending on the polarity sign) to the chosen layer's activations during the forward pass, shifting the residual stream in the targeted Side Feature direction.

\paragraph{DPO}
We train \qwenNineBAbbr, \qwenTwentySevenBAbbr, \llamaThreeOneAbbr, \gptOssTwentyBAbbr, \gemmaEfourBAbbr, and \texttt{Gemma4-31B} with DPO.
We treat the Prompting intervention output (generate with \promptText{"Please be \{main\} but (not) \{side\}"}) as a preference target and styled-only output (generated with \promptText{"Please be \{main\}"}) as a rejected target to train each model within each domain on the $4{,}800$ messages constructed from the training split, with the same initial message and the same per-message length constraint.
The trained checkpoint is then prompted at test time with the same single-style prompt (no prompt mitigation) on the test set. 

\paragraph{SFT}
SFT shares its data construction and training framework with DPO using Prompting intervention output as target output, while we use \promptText{"Please be \{main\}"} in its input prompt at inference time. As with DPO, the trained checkpoint is prompted at test time with the single-style prompt only.

\subsection{Results}
\label{sec:cse_results}

\begin{figure*}[t!]
    \centering
    \includegraphics[width=\linewidth]{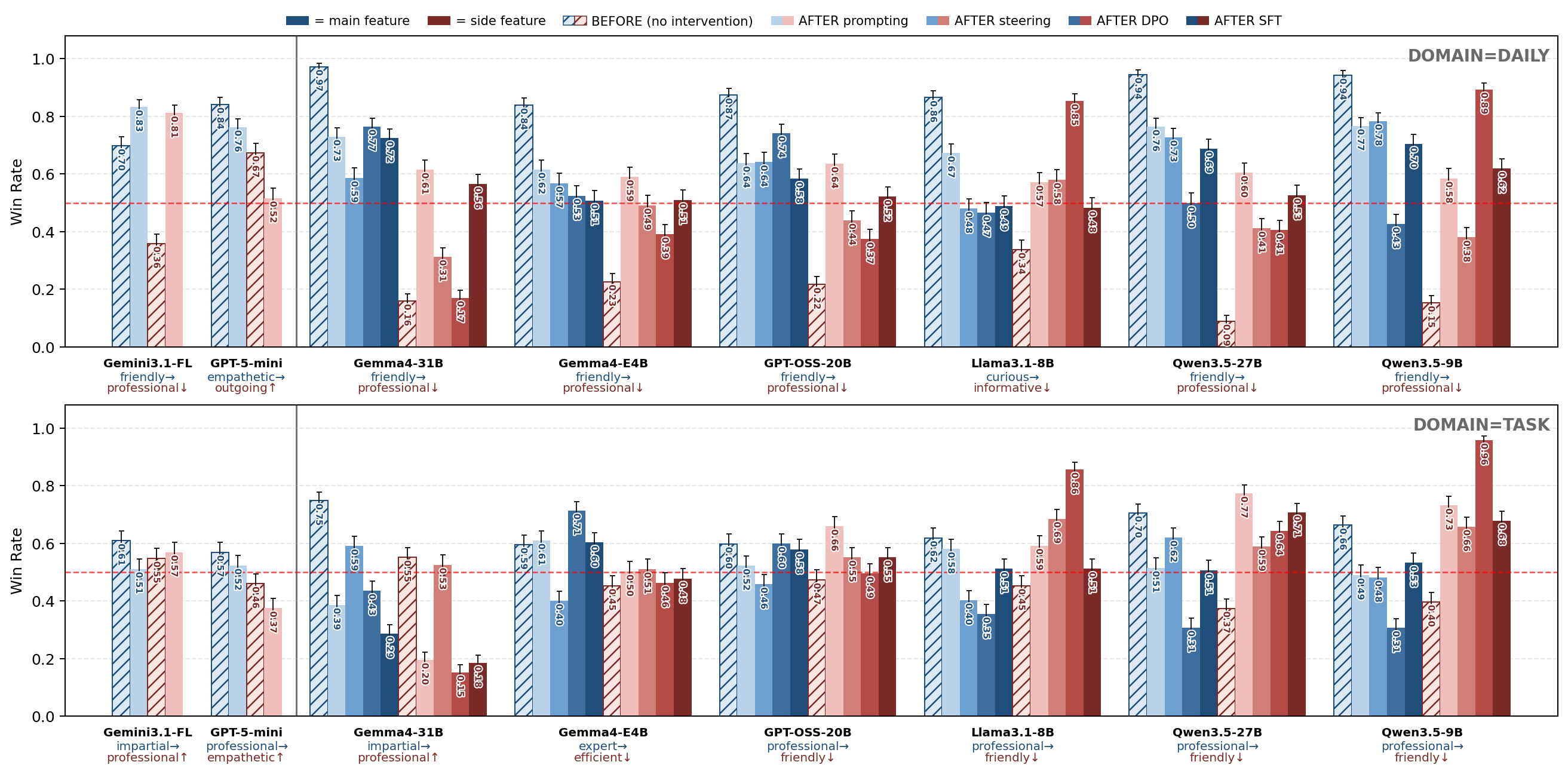}
    \caption{BEFORE/AFTER win-rate proportions \(W\in[0,1]\) for all four interventions (Prompting, Steering, DPO, SFT) across the 16 (model, domain) side effect pairs. Top panel: daily domain. Bottom panel: task domain. Per group: 5 blue bars (Main Feature) then 5 red bars (Side Feature); the hatched bar is BEFORE (no intervention) and the four solid bars darken light~$\rightarrow$~dark in method order (prompting, steering, DPO, SFT). Error bars are 95\% binomial CIs. No-training-only models (\texttt{gemini-flashlite} and \texttt{gpt5mini}) sit on the left of each panel, separated from the trainable open-weight models by a vertical divider. Judge: \texttt{GPT-OSS-120B}.}
    \label{fig:merged_comparison}
\end{figure*}


Figure~\ref{fig:merged_comparison} shows the per-pair before/after win rates, Appendix Table~\ref{tab:MSE_results} provides the exact values, and Table~\ref{tab:mitigation_summary} reports aggregate method-level results. The results reveal a complex landscape of trade-offs between maintaining stylistic adherence and reducing SES-selected Side Feature shifts.

\begin{table}[t]
\centering
\renewcommand{\arraystretch}{1.1}
\resizebox{\columnwidth}{!}{%
\begin{tabular}{lrrcr}
\toprule
\textbf{Method} & \textbf{$\Delta_{\mathrm{Main}}$ (pp, $\uparrow$)} & \textbf{$\Delta_{\mathrm{Side}}$ (pp, $\uparrow$)} & \textbf{Main Flip ($\downarrow$)} & \textbf{Side Flip ($\uparrow$)} \\
\midrule
Prompting & {\boldmath$-13.5$} & $+28.5$              & {\boldmath$12.5\%$} & {\boldmath$87.5\%$} \\
Steering  & $-21.9$            & $+19.2$              & $41.7\%$            & $50.0\%$ \\
DPO       & $-26.9$            & {\boldmath$+29.8$}   & $58.3\%$            & $50.0\%$ \\
SFT       & $-22.1$            & $+26.6$              & $16.7\%$            & $83.3\%$ \\
\bottomrule
\end{tabular}%
}
\caption{Mitigation results. Prompting is aggregated across all 8 generation models (16 model--domain settings), whereas Steering, DPO, and SFT are aggregated across the 6 open-weight models (12 model--domain settings). Header arrows mark the desired direction.}
\label{tab:mitigation_summary}
\end{table}

\paragraph{Effectiveness of Side Effect Restoration}
The four mitigation methods, \textit{Prompting}, \textit{Steering}, \textit{DPO}, and \textit{SFT}, broadly alleviate the side effect relative to the non-intervened styled response, though the strength of restoration varies substantially across methods and side effects, as shown in Figure~\ref{fig:merged_comparison}. For example, on \qwenNineBAbbr{} in the Task domain, the \textit{Professional} prompt originally suppresses \textit{Friendly} to a win rate of 0.40; Prompting restores it to 0.73, Steering to 0.66, DPO to 0.96, and SFT to 0.68. Similarly, on \llamaThreeOneAbbr{} in the same domain, the suppressed \textit{Friendly} score of 0.45 is recovered to 0.59 by Prompting, 0.69 by Steering, 0.86 by DPO, and 0.51 by SFT. However, restoration is not uniform. On \gemmaThirtyOneBAbbr{} in the Daily domain, where the \textit{Friendly} prompt drives \textit{Professional} down to 0.16, Prompting and SFT substantially recover the trait, reaching 0.61 and 0.57 respectively, while Steering (0.31) and DPO (0.17) leave it close to the suppressed baseline.

\paragraph{The Alignment Tax}
Although side effects are generally mitigated, preserving the strength of the \mainFeature{} is more difficult. Figure~\ref{fig:merged_comparison} shows that some \mainFeature{} win rates drop sharply after restoration, while others are maintained or even strengthened. A high-degradation case appears for \llamaThreeOneAbbr{} in the Daily domain: when restoring \textit{Informative} while prompting for \textit{Curious}, the \textit{Curious} win rate falls from 0.87 to 0.67 under Prompting, and further to 0.48, 0.47, and 0.49 under Steering, DPO, and SFT, respectively. This suggests that delivering information directly can conflict with the inquisitive stance required by \textit{Curious}. In contrast, some feature pairs exhibit synergy rather than competition. For \gemmaEfourBAbbr{} in the Task domain, restoring \textit{Efficient} while prompting for \textit{Expert} preserves or improves the \textit{Expert} win rate: compared with the original 0.60, Prompting reaches 0.61, DPO reaches 0.71, and SFT remains at 0.60. This indicates that expertise and efficiency can reinforce one another in task-oriented responses, rather than crowding each other out.

\paragraph{Method-Level Comparison}
Table~\ref{tab:mitigation_summary} summarizes the aggregate performance of the four mitigation methods. \textit{Prompting} achieves the smallest average \mainFeature{} loss ($\Delta_{\mathrm{Main}} = -13.5$ pp), the lowest Main Flip rate ($12.5\%$), and the highest Side Flip rate ($87.5\%$), while remaining competitive in \sideFeature{} restoration ($\Delta_{\mathrm{Side}} = +28.5$ pp). \textit{DPO} obtains the highest average \sideFeature{} gain ($\Delta_{\mathrm{Side}} = +29.8$ pp) and the strongest individual improvements, increasing \textit{Professional} on \qwenNineBAbbr{} from $0.15$ to $0.89$ (Daily) and \textit{Friendly} on \qwenNineBAbbr{} from $0.40$ to $0.96$ (Task). However, DPO also shows the largest average \mainFeature{} decrease ($\Delta_{\mathrm{Main}} = -26.9$ pp) and the highest Main Flip rate ($58.3\%$). \textit{SFT} falls between Prompting and DPO, with $\Delta_{\mathrm{Side}} = +26.6$ pp, $\Delta_{\mathrm{Main}} = -22.1$ pp, and a Side Flip rate of $83.3\%$. \textit{Steering} yields the smallest \sideFeature{} gain ($\Delta_{\mathrm{Side}} = +19.2$ pp), a comparable \mainFeature{} decrease to SFT ($\Delta_{\mathrm{Main}} = -21.9$ pp), and the lowest Side Flip rate ($50.0\%$). Overall, no single intervention dominates every case, but the aggregate results show that natural-language Prompting wins on three of the four metrics, whereas training-based methods trade higher peak \sideFeature{} restoration for larger main-feature degradation.

\section{Discussion}
\label{discussion}

\label{sec:discussion}

\subsection{Side Effects as Bundled Style Realizations}

Using two illustrative Gemma-4-E4B cases, we compare \textit{Styled} vs \textit{Neutral} responses to identify how a single-feature prompt realizes the Main Feature together with a Side Feature shift, then matched \textit{Styled} vs \textit{Prompt-Mitigated} responses to identify what changes when the Side Feature recovers. We focus on keyword and content bundles; Appendix~\ref{app:response_difference_analysis} gives the full descriptive analysis.

\paragraph{Friendly $\rightarrow$ less Professional: warmth cues and interactional extension.} Relative to Neutral responses, Friendly responses add \emph{wonderful}, \emph{great}, and \emph{lovely}, together with reassurance, offers of help, well wishes, and follow-ups. This warmth bundle strengthens Friendliness but moves beyond a restrained Professional register, creating the Side Feature shift. Across the matched comparisons, both conditions draw from overlapping evaluative vocabulary. However, strongly warm words such as \emph{wonderful}, \emph{great}, and \emph{fun} occur more often only in the Styled response. Their absence from the matched mitigated response is associated with higher Professional preference, suggesting that mitigation recovers Professionalism by attenuating overt warmth. Mitigated-only content instead emphasizes direct assistance and concrete task details rather than social extension. This selective attenuation coincides with higher Professional preference and lower Friendly preference.

\paragraph{Expert $\rightarrow$ less Efficient: breadth and answer prioritization.} Relative to Neutral responses, Expert responses combine \emph{specific}, \emph{complex}, \emph{robust}, and depth-oriented vocabulary with mechanisms, alternatives, edge cases, and implementation details. This breadth demonstrates Expertise but can obscure the actionable answer, producing the less-Efficient judgment. At the aggregate level, mitigation shows lexical overlap: both styled-only and mitigated-only lists include \emph{specific}, \emph{complex}, and \emph{robust}. Yet \emph{rigorous}, \emph{deep}, and \emph{critical} are more prominent in the styled-only set, while \emph{efficient}, \emph{operational}, \emph{dependable}, and \emph{direct} are more prominent in the mitigated-only set. Content differences follow the same functional contrast: broader explanatory detail in Styled responses and more immediate requirements and next steps under mitigation. These differences coincides with higher Efficient preference and lower Expert preference.

\paragraph{Implications for Style Control and Mitigation}

Together, the comparisons support \emph{bundled style realizations}: a Main Feature label recruits keywords and content choices that also shift the Side Feature. Mitigated responses remain anchored to the same task, while their aggregate lists overlap even as different cues dominate. The resulting higher Side Feature but lower Main Feature preference motivates controls such as \emph{warm but formal} or answer-first Expertise with optional caveats.



\subsection{Asymmetry of Style Feature Effects}
\label{disc:asymmetry}
When prompting for feature $a$ affects feature $b$ differently from the reverse prompt, the relationship is \emph{asymmetric}. The effect matrix is therefore a directed map of interventions rather than an undirected map of feature associations.

Two recurring reversals illustrate this pattern across models in Appendix Figure~\ref{fig:combined_heatmap}. In Daily dialogues, \textit{Helpful} prompts make responses more \textit{Outgoing}, whereas \textit{Outgoing} prompts make them less \textit{Helpful}. In Task dialogues, \textit{Expert} prompts increase perceived \textit{Curiosity}, whereas \textit{Curious} prompts reduce perceived \textit{Expertise}. One possible interpretation is that utility-focused styles sometimes recruit interactional cues. Helpful responses may include proactive offers and conversational follow-ups that read as Outgoing, while Expert responses may explore assumptions and alternatives that read as Curious. Conversely, explicitly Outgoing or Curious prompts may emphasize sociability or questioning at the expense of direct assistance or authoritative presentation. These interpretations are descriptive hypotheses, not evidence of a causal mechanism.

Asymmetry therefore suggests that style labels invoke cue bundles with unequal scope. Pairwise effects should consequently be treated as ordered: evidence for $a\!\rightarrow\!b$ does not establish $b\!\rightarrow\!a$. Style-control and mitigation methods should evaluate both directions rather than assume compatibility from the labels alone.

\subsection{Mitigation Comes With Trade-off} 
Our mitigation results suggest that intervening on a side effect is not merely a matter of applying a stronger method. Instead, mitigation exposes an \emph{alignment tax}: restoring a suppressed Side Feature often weakens the intended Main Feature that the original style prompt was meant to induce. This pattern is clearest in the contrast between Prompting and DPO. Prompting provides the best aggregate trade-off, with the smallest Main Feature loss, the lowest Main Feature flip rate, and the highest Side Feature flip success rate, while still restoring the Side Feature. DPO, by contrast, achieves the largest Side Feature gain, but at the cost of the largest Main Feature degradation and the highest Main Feature flip rate. These results indicate that a training-based intervention can have the alignment tax. More broadly, mitigation should therefore seek Pareto improvements across both objectives, rather than maximize Side Feature recovery alone.

\section{Conclusion}
\label{sec:conclusion}
This work shows that style prompts for conversational agents can produce systematic, SES-identified side effects. Through a survey of ACL Anthology papers and controlled experiments across Task and Daily dialogues, we find that prompting for one style feature can shift other stylistic dimensions in directions conflicting with the semantic-similarity prior, showing that style features are deeply entangled rather than independent attributes. Our mitigation experiments further show that intervening on side effects is difficult. Prompting, activation steering, DPO, and SFT can often mitigate side effects, but they frequently degrades the \mainFeature{} intended by the original prompt. Therefore, stylistic control should be treated as a multi-objective optimization problem, where reliable mitigation preserves prompted styles while constraining SES-identified side effects.

\section*{Limitations}
\label{sec:limitation}

This study is deliberately scoped to controlled first-response conversational settings. This design is central to our objective: by holding the initiating message, topic frame, response length, and neutral baseline fixed, we can attribute changes in Side Feature judgments to the prompted Main Feature rather than to uncontrolled dialogue-history effects. The same framework naturally extends to longer human-agent interactions, where user adaptation, topic shifts, and accumulated context may reveal how the style-feature dependencies identified here evolve over sustained dialogue.

Our experimental grid is designed to capture style controls that are common in current conversational-agent research. We study 13 English adjective-style features extracted from recent ACL Anthology papers, two complementary dialogue domains, and eight generation models spanning open-weight and API-served systems. This scope enables systematic cross-feature, cross-domain, and cross-model comparison while keeping the full pairwise generation-and-rating sweep tractable. As conversational-agent practice expands to new languages, domains, and style vocabularies, the same survey-to-effect-matrix pipeline can be reused to construct broader maps of stylistic entanglement.

The mitigation experiments compare four representative intervention families: natural-language Prompting, activation Steering, DPO, and SFT. Evaluating these methods under a matched data split and a shared win-rate protocol allows us to characterize the trade-off frontier between restoring a Side Feature and preserving the prompted Main Feature. This setup also motivates extensions toward iterative re-prompting, explicitly Pareto- or multi-objective training, and interventions that address multiple Side Feature shifts simultaneously.

Finally, SES identifies statistically significant empirical shifts whose directions relative to the neutral baseline are opposite to a cosine-similarity prior centred at the off-diagonal mean. The sign-discordance gate focuses the intervention targets on cases that are especially diagnostic of feature entanglement, rather than ranking prior-congruent patterns. Because the cosine-similarity centring is calibrated to the selected feature inventory, SES is a relative ranking tool for this study. Accordingly, SES does not establish whether a selected shift conflicts with developer intent or user expectations, nor whether it is harmful or undesirable in a particular application. Larger or domain-specific style taxonomies can re-estimate the same statistics, or introduce complementary scoring variants, to study additional forms of cross-feature interaction.

We use “intervention” in the narrow experimental sense of changing the system prompt while holding the initiating message, topic frame, response-length target, generator model, and evaluation protocol fixed. Our estimand is the effect of a Main-Feature prompt on the probability that a fixed judge rates the generated response as stronger along a Side-Feature axis than a neutral-prompt response. We do not claim to identify causal relations among latent style constructs themselves, nor do we claim that one style feature universally causes another across all prompts, users, judges, or multi-turn settings.

\bibliography{custom}

\appendix

\section{Details of Style Feature Extraction Survey}
\label{appendix:style_feature_extraction_survey_details}

This section expands the survey methodology in Section~\ref{sec:style_feature_extract} and summarizes how candidate style features were collected, filtered, and interpreted.

\subsection{Survey Collection and Feature Inventory}

\label{appendix:data_collection_pipeline}

\begin{figure*}[!t]
    \centering
    \includegraphics[width=0.8\textwidth]{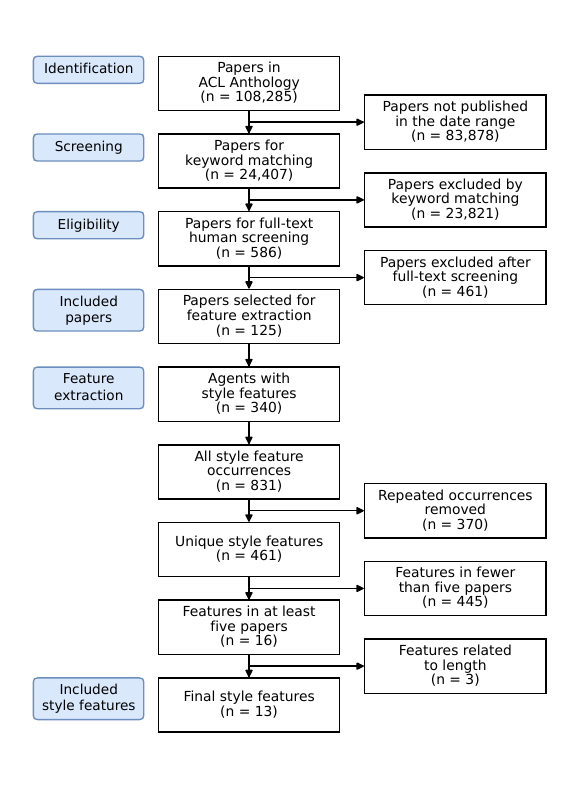}
    \caption{Data collection and screening pipeline following the PRISMA framework for the survey of conversational-agent papers and the selection of style-related features.}
    \label{fig:data_collection}
\end{figure*}

Figure~\ref{fig:data_collection} visualises the PRISMA-based pipeline used to collect the survey of conversational-agent papers and to extract the candidate style features described in Section~\ref{sec:style_feature_extract}.

\label{appendix:feature_distribution}

\begin{figure*}[!t]
    \centering
    \includegraphics[width=0.8\textwidth]{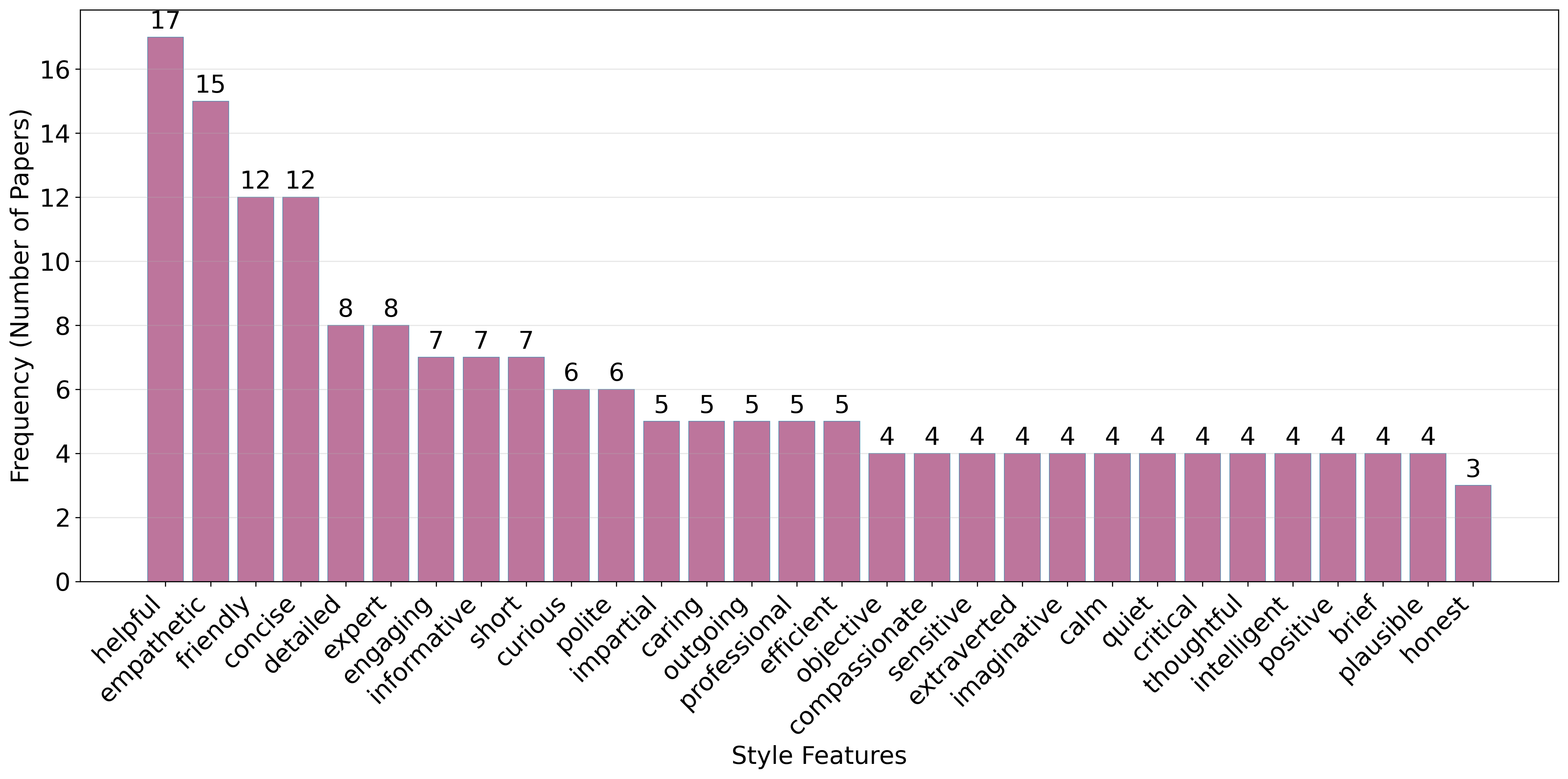}
    \caption{Number of surveyed papers containing each of the 30 most frequent style features.}
    \label{fig:feature_distribution}
\end{figure*}

Figure~\ref{fig:feature_distribution} shows the 30 most frequent style features and the number of surveyed papers in which each appeared.

Table~\ref{tab:style_features_frequency} lists the papers in which each of our 16 filtered style features appears.

\begin{table*}[!t]
\centering
\renewcommand{\arraystretch}{0.85} 

\resizebox{\textwidth}{!}{%
    \footnotesize 
    \begin{tabular}{lcp{12.5cm}}
    \toprule
    \textbf{Style Feature} & \textbf{Freq.} & \textbf{Related Papers} \\
    \midrule
    helpful & 17 & \citet{jiang-etal-2023-superdialseg}, \citet{kwan-etal-2024-mt}, \citet{lee-etal-2025-llms}, \citet{li-etal-2024-cleangen}, \citet{liu-etal-2024-lara}, \citet{meade-etal-2023-using}, \citet{mishra-etal-2023-e}, \citet{rana-etal-2025-zero}, \citet{saha-srihari-2024-integrating}, \citet{stureborg-etal-2024-tailoring}, \citet{wang-etal-2023-cue}, \citet{xu-etal-2024-kiwi}, \citet{xu-etal-2024-safedecoding}, \citet{yang-ettinger-2023-follow}, \citet{zhang-etal-2024-escot}, \citet{zhao-etal-2025-diversity}, \citet{zhou-etal-2024-think} \\
    empathetic & 15 & \citet{almutairi-etal-2024-synthetic}, \citet{cheng-etal-2025-exploring}, \citet{feng-etal-2025-emocharacter}, \citet{finch-etal-2023-leveraging}, \citet{li-etal-2024-eden}, \citet{mishra-etal-2023-e}, \citet{njifenjou-etal-2025-enabling}, \citet{nozue-etal-2024-multimodal}, \citet{reguera-gomez-etal-2025-empathy}, \citet{svikhnushina-pu-2023-approximating}, \citet{wang-etal-2023-cue}, \citet{wang-etal-2025-sibyl}, \citet{zhang-etal-2024-escot}, \citet{zhang-etal-2024-stickerconv}, \citet{zhou-etal-2024-think} \\
    concise & 12 & \citet{afzal-etal-2024-towards}, \citet{agrawal-etal-2024-dialog}, \citet{chalamalasetti-etal-2023-clembench}, \citet{feng-etal-2024-continual}, \citet{li-etal-2024-curriculum}, \citet{movva-etal-2024-annotation}, \citet{siyan-etal-2024-using}, \citet{sreedhar-etal-2024-canttalkaboutthis}, \citet{taranukhin-etal-2024-empowering}, \citet{wang-etal-2023-target}, \citet{wang-etal-2024-book2dial}, \citet{wang-etal-2025-sibyl} \\
    friendly & 12 & \citet{dou-etal-2023-plugmed}, \citet{lechner-etal-2023-challenges}, \citet{lee-etal-2023-prompted}, \citet{liu-etal-2025-cape}, \citet{njifenjou-etal-2025-enabling}, \citet{reddy-etal-2023-social}, \citet{reguera-gomez-etal-2025-empathy}, \citet{semnani-etal-2023-wikichat}, \citet{seo-lee-2024-diagesc}, \citet{tsubota-kano-2024-text}, \citet{wang-etal-2023-cue}, \citet{zhang-etal-2024-escot} \\
    detailed & 8 & \citet{deng-etal-2024-wildvis}, \citet{jang-etal-2024-mixed}, \citet{kirstein-etal-2024-tell}, \citet{li-etal-2024-cleangen}, \citet{liu-etal-2024-lara}, \citet{sobhani-etal-2025-language}, \citet{sreedhar-etal-2024-canttalkaboutthis}, \citet{zhao-etal-2025-diversity} \\
    expert & 8 & \citet{ferron-etal-2023-meep}, \citet{lee-etal-2025-aman}, \citet{ou-etal-2024-dialogbench}, \citet{rachidi-etal-2025-design}, \citet{rana-etal-2025-zero}, \citet{saley-etal-2024-meditod}, \citet{wevelsiep-etal-2025-voice}, \citet{yang-etal-2024-speaker} \\
    engaging & 7 & \citet{almutairi-etal-2024-synthetic}, \citet{finch-choi-2025-leveraging}, \citet{jiang-etal-2024-unknown}, \citet{lee-etal-2023-prompted}, \citet{phukan-etal-2024-ecis}, \citet{wang-etal-2023-target}, \citet{yu-etal-2024-llms} \\
    informative & 7 & \citet{almutairi-etal-2024-synthetic}, \citet{jiang-etal-2024-unknown}, \citet{raposo-etal-2023-prompting}, \citet{siro-etal-2024-context}, \citet{sreedhar-etal-2024-canttalkaboutthis}, \citet{wang-etal-2023-target}, \citet{yu-etal-2024-llms} \\
    short & 7 & \citet{brown-etal-2024-generation}, \citet{jandaghi-etal-2024-faithful}, \citet{li-etal-2024-curriculum}, \citet{reddy-etal-2023-social}, \citet{sreedhar-etal-2024-canttalkaboutthis}, \citet{sun-etal-2024-conscendi}, \citet{vath-etal-2024-towards} \\
    curious & 6 & \citet{lee-etal-2025-llms}, \citet{murakhovska-etal-2023-salespeople}, \citet{njifenjou-etal-2025-enabling}, \citet{petrak-etal-2024-learning}, \citet{wang-etal-2023-cue}, \citet{wang-etal-2023-target} \\
    polite & 6 & \citet{abdullin-etal-2023-synthetic}, \citet{afzal-etal-2024-towards}, \citet{lee-etal-2023-prompted}, \citet{li-etal-2024-cleangen}, \citet{liu-etal-2024-lara}, \citet{petrak-etal-2024-learning} \\
    caring & 5 & \citet{almutairi-etal-2024-synthetic}, \citet{fei-etal-2024-empathyear}, \citet{feng-etal-2025-emocharacter}, \citet{wang-etal-2023-cue}, \citet{zhu-etal-2025-integrating} \\
    efficient & 5 & \citet{lee-etal-2025-llms}, \citet{liu-etal-2025-cape}, \citet{niu-etal-2024-enhancing}, \citet{wang-etal-2023-target}, \citet{zhu-etal-2025-integrating} \\
    impartial & 5 & \citet{duan-etal-2025-guidellm}, \citet{madani-etal-2025-recipe}, \citet{meade-etal-2023-using}, \citet{raju-etal-2024-constructing}, \citet{zhao-etal-2025-diversity} \\
    outgoing & 5 & \citet{he-etal-2025-madial}, \citet{lee-etal-2025-llms}, \citet{liu-etal-2025-cape}, \citet{wang-etal-2023-cue}, \citet{wang-etal-2023-target} \\
    professional & 5 & \citet{almutairi-etal-2024-synthetic}, \citet{kirstein-etal-2024-tell}, \citet{oshima-etal-2024-gap}, \citet{rachidi-etal-2025-design}, \citet{saley-etal-2024-meditod} \\
    \bottomrule
    \end{tabular}%
}
\caption{Overview of the 16 identified style features appearing in at least 5 papers, with references of all papers which contain those features. All terms have been adjectivized and deduplicated for consistency.}
\label{tab:style_features_frequency}
\end{table*}

\subsection{Feature Distribution and Survey Findings}
\label{appendix:feature_extraction_insights}

The survey vocabulary is strongly right-skewed. \textit{Helpful} ($N=17$) and \textit{empathetic} ($N=15$) are the most frequent descriptors, followed by \textit{concise} and \textit{friendly} ($N=12$ each). Of the $461$ unique extracted terms, $349$ ($75.7\%$) appear in only one paper. The 16-feature candidate inventory spans both functional constraints (e.g., \textit{concise} and \textit{detailed}) and interpersonal styles (e.g., \textit{empathetic} and \textit{outgoing}). Separate occurrences of near-synonyms such as \textit{concise}, \textit{short}, and \textit{brief} also indicate that the literature does not use a fully standardized style taxonomy.

\begin{figure}
    \centering
    \includegraphics[width=\linewidth]{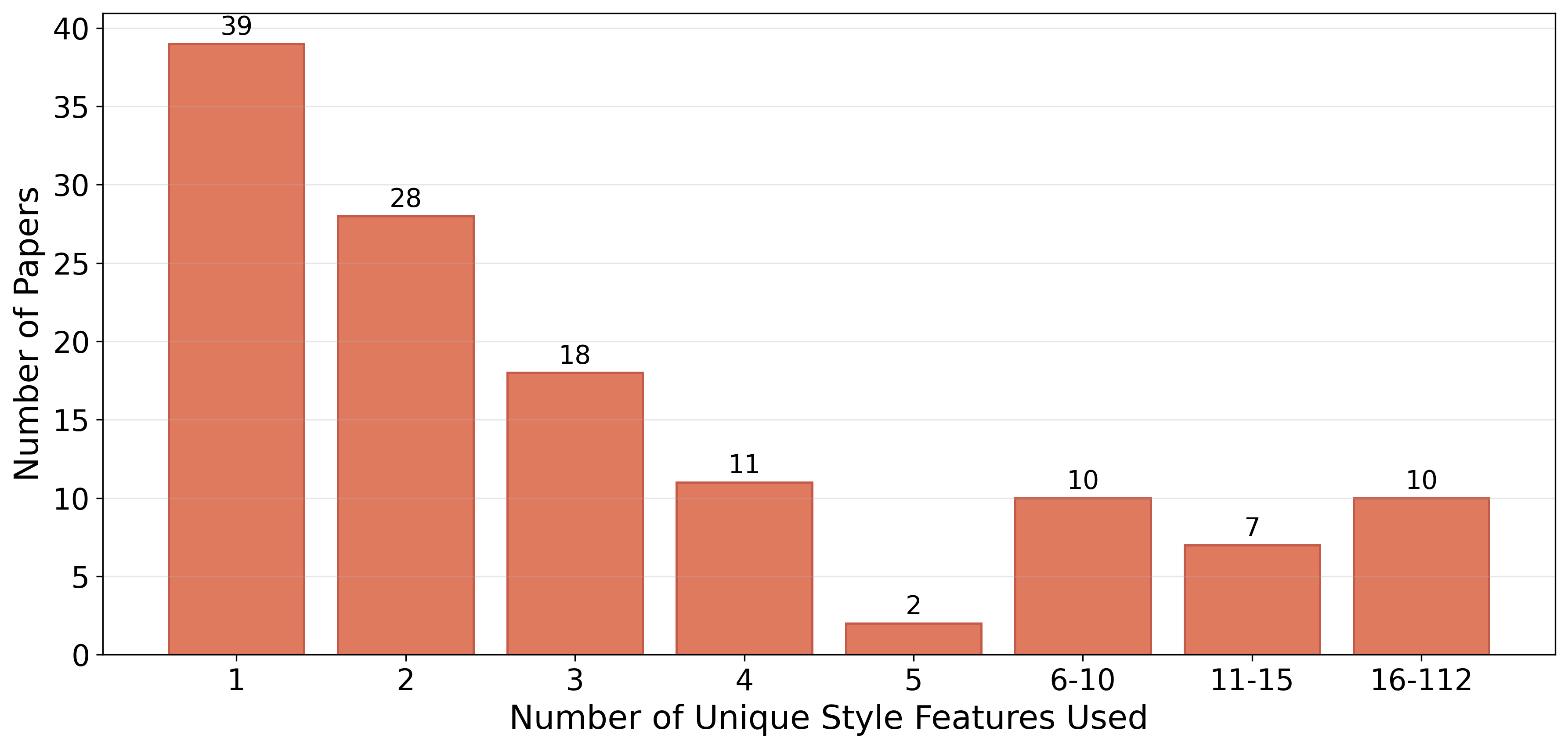}
    \caption{Distribution of papers categorized by the number of unique style features used. The distribution highlights a polarization between minimalist designs (single-feature) and broad-spectrum evaluations (16+ features).}
    \label{fig:paper_feature_count}
\end{figure}

Style control is usually specified with a small number of terms. Among the $340$ surveyed agents, the median is one feature and $176$ agents ($51.8\%$) use a single feature. At the paper level, Figure~\ref{fig:paper_feature_count} shows that most studies use only a few style descriptors, while a small tail evaluates substantially larger inventories.

\section{Details of Data Generation and Annotation}
\label{appendix:data_generation_annotation_details}

This section provides additional details on the datasets, topic construction, response generation, length control, and model-based annotation procedures described in Section~\ref{sec:generation_pipeline}.

\subsection{Dataset Construction and Message Sampling}
\label{sec:selection_of_topics}

\label{sec:dataset_topics_annotation}

We use LMSYS-Chat-1M for task-oriented inquiries and DailyDialog for casual, daily-life interactions. For each domain, we construct a fixed taxonomy of 10 topics; Appendix Table~\ref{tab:domain_topic_examples} lists the topics and representative initial messages. The annotation and sampling pipeline produces two disjoint pools per topic: 10 initial messages for side effect identification and 80 for mitigation experiments, for 90 unique initial messages per topic in total.

This section details (i) how user prompts are annotated into the task-domain and daily-domain taxonomies and (ii) how the resulting label pools are sampled into the two experimental sets.

\begin{table*}[!t]
\centering
\small
\renewcommand{\arraystretch}{1.15}
\begin{tabular}{c|p{5cm}|p{7cm}}
\hline
\textbf{Domain} & \textbf{Topic} & \textbf{Initial Message Examples} \\
\hline

\multirow{22}{*}{\shortstack{\textbf{LMSYS-Chat-1M}\\ \textbf{(Task)}}}
 & Answering questions based on passages & Who was the first man on the moon? \\
 \cline{2-3}
 & Discussing software errors and solutions & Describe and explain the benefits of OpenBSD \\
 \cline{2-3}
 & Inquiries about specific plant growth conditions & how do I get rid of mosquitos? \\
 \cline{2-3}
 & Requesting introductions for various chemical companies & Explain "reductive amination". \\
 \cline{2-3}
 & Inquiries about AI tools, software design, and programming & write python function to reverse a sentence \\
 \cline{2-3}
 & Text processing (Merged) & sup peeps. Wanna help me with summarizing lyrics? \\
 \cline{2-3}
 & Role-playing scenarios and character interactions (Merged) & Write me an extremely funny tale about pillow hoarding ducks at a luxurious hotel. \\
 \cline{2-3}
 & Geography, travel, and global cultural inquiries & explain why we have traffic lights? \\
 \cline{2-3}
 & Discussing and describing various characters & Tell me something about Stephen King's Dark Tower. \\
 \cline{2-3}
 & Creating and improving business strategies and products & how to make a game project success \\
\hline

\multirow{10}{*}{\shortstack{\textbf{DailyDialog}\\ \textbf{(Daily)}}}
 & Tourism & That is the most beautiful sunset ! \\
 \cline{2-3}
 & Work & Hey , Zina . You're here early today . \\
 \cline{2-3}
 & Politics & Every country should face the history . \\
 \cline{2-3}
 & Finance & It's all over . I'm bankrupt . \\
 \cline{2-3}
 & Health & Are you feeling better today , Bill ? \\
 \cline{2-3}
 & School Life & What can I help you with today ? \\
 \cline{2-3}
 & Attitude \& Emotion & Do you hear what happened to Sally ? \\
 \cline{2-3}
 & Ordinary Life & Excuse me . Is this seat taken ? \\
 \cline{2-3}
 & Culture \& Education & Harry , do you like the opera ? \\
 \cline{2-3}
 & Relationship & I wonder how Sarah and Mat are . \\
\hline
\end{tabular}
\caption{Example Initial Messages by Domain and Topic (10 Topics Per Domain). Task topics are curated by merging overlapping LMSYS clusters; Daily topics follow the DailyDialog taxonomy.}
\label{tab:domain_topic_examples}
\end{table*}

\paragraph{Source Datasets}

For the task domain we draw on two LMSYS-released resources. \textbf{LMSYS-Chat-1M} \citep{zheng2023lmsyschat1m} provides one million conversations collected from the Chatbot Arena interface across $25$ open-weights and proprietary models; in the paper, the authors cluster the first user prompts into $20$ representative topic categories (e.g.\ \textit{Python coding assistance}, \textit{role-playing characters}, \textit{Geography/travel inquiries}). We use both (a) the original \texttt{lmsys/lmsys-chat-1m} corpus for taxonomy curation and the larger 80-per-topic extension, and (b) the smaller, paired-completion \texttt{lmsys/chatbot\_arena\_conversations} corpus for the original 10-per-topic seed set.
For the daily domain we use \textbf{DailyDialog} \citep{li-etal-2017-dailydialog}, a manually labeled English chitchat corpus of $13{,}118$ multi-turn dialogues. The paper defines $10$ daily-life topics together with their per-dialogue proportions (Table~4 of \citet{li-etal-2017-dailydialog}). The dataset's original distribution site (\texttt{yanran.li}) is no longer reachable; we obtain dialogue text via the community mirror \href{https://huggingface.co/datasets/roskoN/dailydialog}{on Hugging Face}, which preserves the dialogue, dialogue-act, and emotion files but does \emph{not} preserve the dialogue-level topic labels. We therefore re-annotate topics from scratch via \texttt{gpt-4o-mini}, as described below.

\paragraph{Curating a Task-Domain Taxonomy}

The LMSYS-Chat-1M paper's $20$ inferred clusters contain semantic overlaps (e.g.\ separate clusters for \textit{Python coding assistance} vs.\ \textit{Python coding examples}; separate clusters for \textit{Role-playing characters} vs.\ \textit{Role-playing AI chatbot capabilities}; multiple clusters covering text manipulation). To obtain a tractable label space we manually merged near-duplicate clusters and dropped categories whose surface form was either redundant or ill-suited as a stylistic conversation seed (NSFW role-play, SQL queries, factual-consistency probing, sentence-generation for job roles). The final task taxonomy contains the following $10$ topics, listed verbatim as used by all downstream annotation calls:

\begin{itemize}\itemsep0pt
\item Discussing software errors and solutions
\item Inquiries about AI tools, software design, and programming
\item Geography, travel, and global cultural inquiries
\item Text processing — translation, summarization, rewriting, elaboration, and factual-consistency checks
\item Creating and improving business strategies and products
\item Role-playing scenarios and character/chatbot interactions
\item Answering questions based on passages
\item Discussing and describing various characters
\item Inquiries about specific plant growth conditions
\item Requesting introductions for various chemical companies
\end{itemize}

For the daily domain we adopt the $10$ DailyDialog topics directly from the paper: \textit{Ordinary Life}, \textit{School Life}, \textit{Culture \& Education}, \textit{Attitude \& Emotion}, \textit{Relationship}, \textit{Tourism}, \textit{Health}, \textit{Work}, \textit{Politics}, \textit{Finance}.

\paragraph{Annotation Pipeline}

We apply the following annotation procedures in the two domains and run an additional task-domain robustness classification.

\paragraph{Task-domain annotation (\texttt{LMSYS-Chat-1M}, GPT-4o-mini).} We iterate over the conversations in the dataset and extract the first user message of each. Each initial message is passed to \texttt{gpt-4o-mini} \citep{openai_gpt4o_mini_2024} with an $11$-way single-label classification prompt: the $10$ task topics above followed by a catch-all \textit{Other} bucket. The model is instructed to respond with a single integer in $\{1,\dots,11\}$; light post-processing recovers cases where the model emits a prefix such as "Domain $k$" before the digit. Each labeled message is appended to a thread-safe checkpoint CSV (\texttt{task\_categorizations\_checkpoint.csv}) immediately after the API call so that no completed annotation is ever lost across restarts, and a $32$-way parallel pool drives the call rate. This produced $25{,}650$ labeled first-message records in our snapshot.

\paragraph{Daily-domain annotation (DailyDialog via \texttt{roskoN/dailydialog}, GPT-4o-mini).} Because the upstream topic labels are no longer accessible, we re-derive them with the same protocol as the task-domain annotator. We download the train, validation, and test zip archives directly from the HuggingFace mirror, parse the \texttt{dialogues\_\{split\}.txt} files, split each line on the dataset's \texttt{\_\_eou\_\_} delimiter, and retain the dialogue's first utterance as the seed message. To keep the classifier signal strong --- DailyDialog openers are often short greetings ("Hi.", "Excuse me.") that carry little topical content on their own --- the \emph{full} dialogue text (truncated to $2{,}000$ characters) is passed to \texttt{gpt-4o-mini} with the same $11$-way prompt over the $10$ DailyDialog topics + \textit{Other}, while only the first utterance is stored as the eventual seed. Annotations are written incrementally to a resumable checkpoint (\texttt{daily\_categorizations\_checkpoint.csv}) with a thread-safe append guarded by a single header initialization.

\paragraph{Direct classification of LMSYS-Chat-1M into the paper taxonomy (GPT-5-mini).} As a robustness check on the task-domain taxonomy --- and as the source pool for the $80$-per-topic extension described below --- we additionally classify \texttt{lmsys/lmsys-chat-1m} end-to-end: filter to English first messages with $32$--$1{,}536$ characters and deduplicate, yielding $401{,}746$ prompts; then use \texttt{gpt-5-mini} as a $21$-way text classifier that assigns each prompt to one of the LMSYS-Chat-1M paper's $20$ verbatim cluster labels \cite{zheng2023lmsyschat1m} plus an \textit{Other} bucket. The categories are the paper's published labels; \texttt{gpt-5-mini} reads each prompt's raw text and returns a category integer, with no embedding or re-clustering step. The resulting label set contains $198{,}501$ in-taxonomy assignments after dropping \textit{Other}, which serves as the pool the $80$-per-topic extension samples from after mapping the $20$ paper categories onto our $10$ project task topics.

\paragraph{Sampling 10 Initial Messages per Topic (Side Effect Identification)}
\label{sec:sampling_10_initial}

The side effect generation experiments require a fixed, balanced pool of \emph{seed} initial messages, one set of $10$ per topic, so that every reported win rate is computed over a comparable substrate of conversational contexts.

\paragraph{Task domain.} We group all messages by their assigned topic, drop the \textit{Other} bucket, deduplicate exact-string repeats, and stream candidates in dataset order. For each topic we accept a candidate if (a) it is between $32$ and $512$ characters long, (b) it is not a near-duplicate of an already-accepted candidate within the topic (case-folded, whitespace-collapsed comparison), and (c) it is not a malformed or empty turn. We stop accepting once $10$ candidates have been retained for that topic. 

\paragraph{Daily domain.} The same procedure is applied to the daily-domain checkpoint, restricted to the $10$ DailyDialog topics (\textit{Other} dropped), again retaining $10$ unique messages per topic and assigning \texttt{domain} = \texttt{daily}.

\paragraph{Resulting seed set.} The final file \texttt{unique\_first\_messages.csv} contains exactly $200$ rows: $10$ task topics $\times$ $10$ messages plus $10$ daily topics $\times$ $10$ messages. Every conversation in the side effect generation pipeline (Section~\ref{sec:generation_pipeline}, all model families, all style features, all repetitions) is seeded from one of these $200$ initial messages, so the win-rate and SES estimates reported in Sections~\ref{sec:identifying_side_effects}--\ref{sec:mitigation} are matched-conversation comparisons over the same conversational substrate.

\paragraph{Extension to 80 Initial Messages per Topic (Intervention Experiments)}
\label{sec:sampling_80_initial}

The mitigation and intervention experiments require a substantially larger pool of seed messages so that train, validation, and test splits per topic can support DPO and SFT training in addition to evaluation. We extend the seed set to $80$ initial messages per topic in both domains, drawing from larger upstream pools and guaranteeing zero overlap with the $200$-message identification set introduced in Section~\ref{sec:sampling_10_initial}.

\paragraph{Task domain.} The task-domain extension is drawn from the GPT-5-mini classification pool described above (LMSYS-Chat-1M first messages already assigned to one of the LMSYS-paper's $20$ verbatim cluster labels plus \textit{Other}). We construct a fixed many-to-one mapping from the $20$ paper-verbatim labels onto our $10$ project task topics: the two Python-coding clusters and the AI-tools/software-design cluster fold together under \textit{AI tools/programming}; the translation, summarization, and factual-consistency clusters fold together under \textit{Text processing}; the two role-playing clusters fold together under \textit{Role-playing scenarios}; the remaining seven project topics map one-to-one from a single paper cluster. NSFW, toxic-behavior, SQL, and \textit{Other} clusters are excluded by design. From this pool we sample $80$ messages per project topic deterministically (fixed random seed, English-only via a standard language-detection filter, with a $10$--$2{,}000$ character window), excluding any message whose normalized form already appears in the identification seed set.

\paragraph{Daily domain.} The daily-domain extension uses the GPT-4o-mini DailyDialog re-annotation described above. After deduplicating by first utterance, applying the same English and length filters, and classifying each remaining dialogue against the $11$-way DailyDialog taxonomy, we sample $80$ messages per project topic using the same dedup-against-seed protocol as the task domain.

\paragraph{Overlap and uniqueness guarantees.} For both domains the extension is constructed to satisfy two invariants: (i) zero overlap with the $200$-message identification seed set from Section~\ref{sec:sampling_10_initial}, and (ii) no within-extension cross-topic collisions --- a message that has already been assigned to one project topic cannot reappear under any other project topic. The first invariant is enforced by accumulating a normalized-message dedup set from the identification seed before sampling; the second is enforced by sharing that dedup set across topics during the per-topic sampling pass. The resulting extension is therefore additively combinable with the identification seed at downstream consumption time without further deduplication.

\subsection{Generation and Evaluation Protocols}
\label{appendix:generation_evaluation_protocols}

The key prompt templates used in this project are given below.
\label{appendix:prompt_examples}

\paragraph{LLM-as-a-Judge Prompt Template}
\label{evaluator_prompt}

\textbf{System:} \promptText{You are an expert evaluator of conversation quality.}

\noindent \textbf{User:} \promptText{Compare the following two assistant responses and determine which one is more \{feature\}.
Response A:
\{response\_a\}
Response B:
\{response\_b\}
You must choose one. Which response is more \{feature\}? Answer with ONLY "A" or "B" (no ties allowed).
Answer:}

where ``\{feature\}'' refers to one of the 13 evaluated style features.

\paragraph{System Prompt Template for Message Generation With and Without Style Features}
\label{user_assistant_system_prompt}

Neutral: \promptText{"You are having a conversation about \{topic\}."}

\noindent Styled: \promptText{"You are having a conversation about \{topic\}. Please be \{style\_feature\} in your response."}

\label{appendix:length_distribution}

\begin{figure*}[!t]
    \centering
    \includegraphics[width=0.92\linewidth]{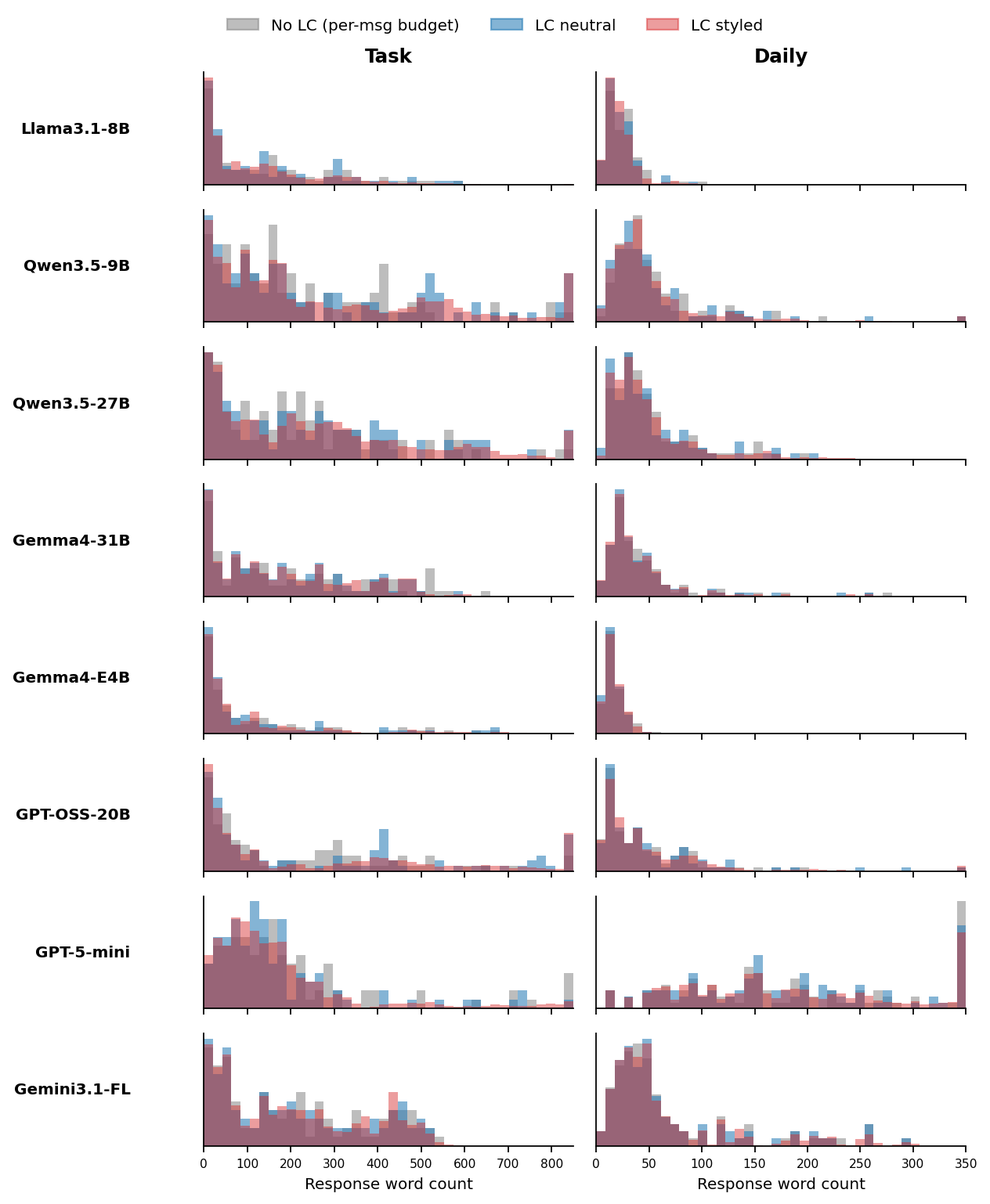}
    \caption{Response-length distributions on the side effect identification initial messages for each of the 8 generation models (rows) in each domain (columns: \textit{Task} left, \textit{Daily} right). Each panel overlays three histograms: \textbf{gray} -- per-message length budget (rounded mean of unconstrained reference generations, $n{=}100$ per panel, used as the no-length-control stand-in); \textbf{blue} -- length-controlled \textit{Neutral} baseline ($n{=}100$ per panel, \texttt{original\_word\_count} from \texttt{rated\_unified.csv}); \textbf{red} -- length-controlled \textit{Styled} responses pooled over all 13 features ($n{=}6{,}500$ per panel, \texttt{styled\_word\_count} from \texttt{rated\_unified.csv}). All histograms are area-normalised; values above the column's $98$th percentile of the styled distribution are clipped into the rightmost bin so the body of each distribution is visible.}
    \label{fig:length_distribution}
\end{figure*}

Figure~\ref{fig:length_distribution} visualises, per (model, domain) setting of the side effect identification phase, the response-length distributions under three regimes:
(i)~the per-message length budget used to set the cap (gray; rounded mean of unconstrained reference generations per initial message; the raw reference outputs were not retained for $7/8$ models, so the budget integer is the closest available stand-in for the no-length-control distribution);
(ii)~the length-controlled \textit{Neutral} response (blue; \texttt{original\_word\_count} from \texttt{rated\_unified.csv}, one per initial message --- the same response the judge saw on the Neutral side); and
(iii)~the length-controlled \textit{Styled} responses, pooled across all 13 features (red; \texttt{styled\_word\_count} from \texttt{rated\_unified.csv}, five per (first-message, feature) --- the same responses the judge saw on the Styled side). For \texttt{GPT-OSS-20B} these word counts come from responses that were already harmony-stripped at rating time, so they reflect only the user-visible answer that contributed to the win-rate measurements.

Across all 16 (model, domain) settings, the length-controlled \textit{Neutral} and \textit{Styled} distributions occupy similar regions around the per-message budgets. This supports the conclusion that the procedure reduces systematic response-length differences between the conditions used for the win-rate measurements in Figure~\ref{fig:correlation_matrix} and Appendix Figure~\ref{fig:combined_heatmap}; it does not establish the absence of all length-related confounding.

\paragraph{Cross-Judge Sensitivity Analysis}
\label{appendix:model_annotation_agreement}

We assess judge sensitivity by re-annotating fixed styled--Neutral response
pairs with the same feature-specific binary-choice question.  Response order
is randomized to avoid position bias, and the response texts remain fixed, so
the comparison isolates annotation variation rather than regeneration.  The
primary annotations use \gptOssOneTwentyB{}, and \texttt{GPT-5.5} fully
re-annotated the outputs from five of the eight generation models.  We report exact
agreement and Cohen's $\kappa$ on decisive pair-level labels, plus Pearson $r_P$ and
Spearman $\rho_S$ correlations between aggregated (\mainFeature{}, \evaluationFeature{})
win-rate cells.

Table~\ref{tab:gpt55_gptoss_agreement} summarizes agreement by generation
model. Across both domains, \texttt{GPT-5.5} and \gptOssOneTwentyB{} agree on
$76.16\%$ of $844{,}154$ annotations ($\kappa=0.521$).  Their pooled matrices
have Pearson $r_P=0.979$, Spearman $\rho_S=0.976$ over $845$ cells, and a mean
absolute difference of $3.37$ percentage points.

\begin{table}[t]
\centering
\small
\setlength{\tabcolsep}{3.0pt}
\resizebox{\columnwidth}{!}{%
\begin{tabular}{lrrrrr}
\toprule
Generator & $N$ & Exact (\%) & $\kappa$ & Pearson $r_P$ & Spearman $\rho_S$ \\
\midrule
\llamaThreeOneAbbr{}      & $168{,}898$ & $74.78$ & $0.494$ & $0.966$ & $0.961$ \\
\qwenTwentySevenBAbbr{}   & $168{,}840$ & $78.90$ & $0.577$ & $0.985$ & $0.986$ \\
\gemmaThirtyOneBAbbr{}    & $168{,}801$ & $78.25$ & $0.564$ & $0.984$ & $0.982$ \\
\gptFiveMiniAbbr{}        & $168{,}883$ & $70.38$ & $0.400$ & $0.959$ & $0.949$ \\
\geminiFlashLiteAbbr{}    & $168{,}732$ & $78.50$ & $0.568$ & $0.982$ & $0.982$ \\
\midrule
All                         & $844{,}154$ & $76.16$ & $0.521$ & $0.979$ & $0.976$ \\
\bottomrule
\end{tabular}%
}
\caption{Agreement between \texttt{GPT-5.5} and \gptOssOneTwentyB{} on fixed
response pairs. Exact is a percentage; Pearson $r_P$ and Spearman $\rho_S$ compare the
combined-domain $13\times13$ win-rate matrices (169 cells per generator; 845
pooled).}
\label{tab:gpt55_gptoss_agreement}
\end{table}

Pair-level $\kappa$ is moderate, whereas high matrix correlations indicate
greater robustness after aggregation.  Cross-LLM agreement is a sensitivity
check, not human construct validation.

\section{Details of Style Feature Effect Analysis}
\label{appendix:style_feature_effect_analysis_details}

This section presents the complete style feature effect matrices underlying Section~\ref{correlation results}.


\subsection{Full Win-Rate Matrices}
\label{appendix:full_win_rate_matrix_each_domain_model}

\begin{figure*}[p]
    \centering
    \includegraphics[height=0.86\textheight,keepaspectratio]{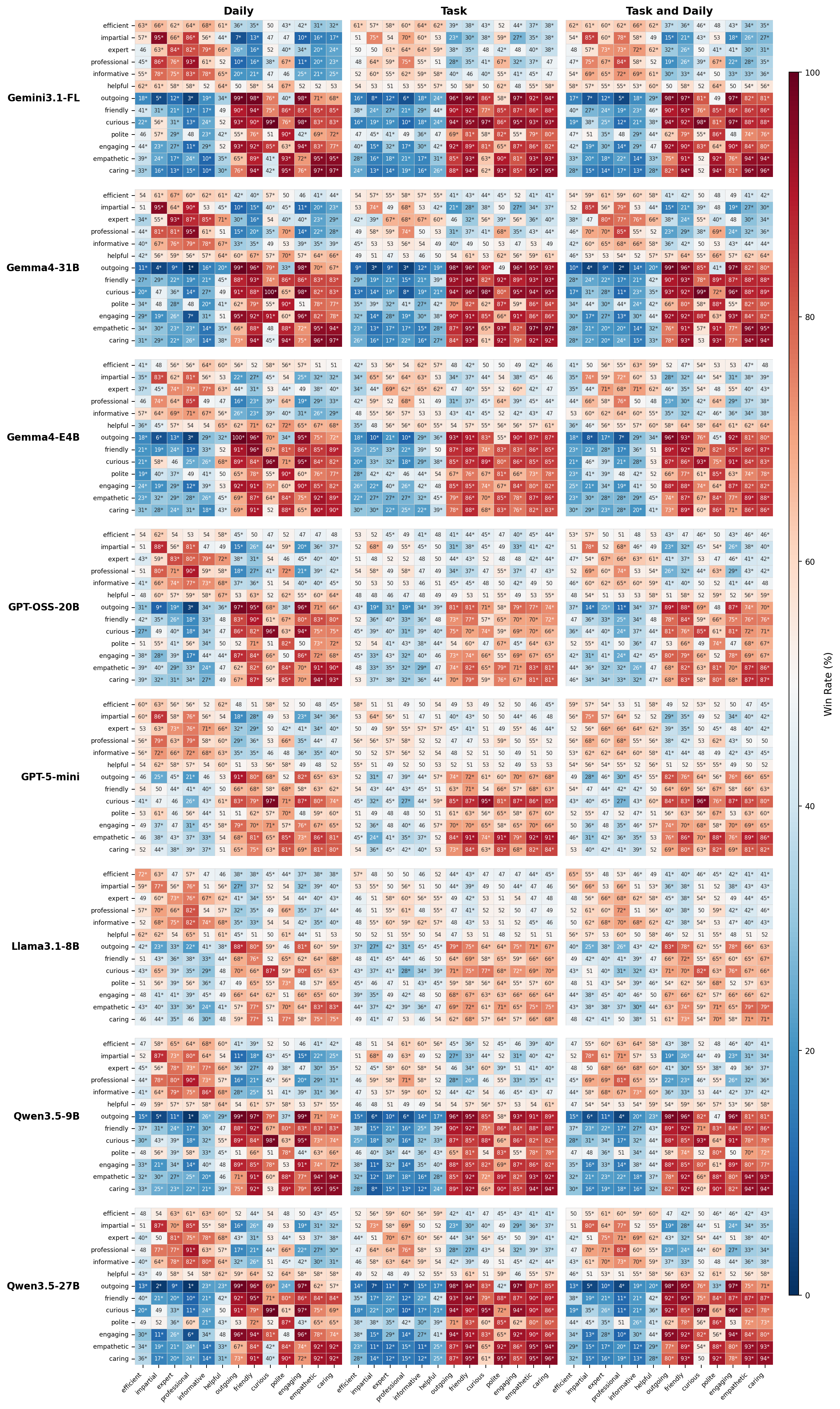}
    \captionof{figure}{Win-rate effect matrices for all 8 generation models across the \textit{Daily}, \textit{Task}, and \textit{Task and Daily} (both domains pooled) splits. Each subplot is a $13 \times 13$ heatmap: rows are prompted \mainFeature{}s, columns are evaluated \evaluationFeature{}s. Each entry reports integer-rounded \(100W(a,b)\%\) for the styled response over the neutral baseline; * marks \(p_W(a,b) \le 0.05\) (binomial). Red $>$ 50\% / blue $<$ 50\%. Shared colorbar on the right.}
    \label{fig:combined_heatmap}
\end{figure*}

Figure~\ref{fig:combined_heatmap} shows the win rates across all models and domains, providing the complete empirical matrices underlying the main-text analyses.

\begingroup
\raggedbottom

\subsection{Magnitude, Prevalence, and Semantic Similarity}
\label{appendix:feature_independence_quantification}

We quantify cross-feature effects over all $13\times12=156$ ordered off-diagonal feature pairs in each of the 16 model--domain matrices (eight models, two domains). Unlike SES-based target selection, this analysis includes every off-diagonal pair, without filtering by feature category or SES. Table~\ref{tab:feature_independence_quantification} reports the mean and range of the resulting matrix-level statistics.

\begin{table}[H]
\centering
\small
\setlength{\tabcolsep}{3pt}

\begin{tabular}{@{}>{\raggedright\arraybackslash}p{0.50\linewidth}rr@{}}
\toprule
Metric & Mean & Range \\
\midrule
Absolute deviation from $50\%$ (pp) & 16.6 & 7.8--22.3 \\
Pairs with $\geq10$ pp deviations & $59.1\%$ & $28.2$--$78.2\%$ \\
Pairs with $\geq10$ pp, Holm-significant deviations & $55.8\%$ & $26.3$--$73.1\%$ \\
Variance explained by similarity ($R^2$) & $12.9\%$ & $8.2$--$17.1\%$ \\
\bottomrule
\end{tabular}
\caption{Magnitude and prevalence of cross-feature effects across the 16 model--domain matrices. Means equally weight the matrices; ranges are minima and maxima. Significance uses one-sample $t$-tests on message-level mean judgments, with Holm correction within each matrix.}
\label{tab:feature_independence_quantification}
\end{table}

\paragraph{Calculation procedure.}
For each off-diagonal pair, we calculate $D_{g,d}(a,b)=100|W_{g,d}(a,b)-0.5|$, expressed in percentage points. Within each matrix, we report the mean of these deviations and the percentage of pairs with $D_{g,d}(a,b)\geq10$.

For significance testing, we first average repeated judgments within each initiating message and then perform a one-sample $t$-test against $0.5$ across the message-level means. Holm correction is applied to all 156 off-diagonal tests within each matrix. The third statistic is the percentage of all 156 pairs satisfying both $D_{g,d}(a,b)\geq10$ and adjusted $p<0.05$.

Separately, we fit an ordinary least-squares regression with an intercept over the 156 pairs:
\[
W_{g,d}(a,b)=\alpha_{g,d}+\beta_{g,d}C(a,b)+\epsilon_{g,d}(a,b),
\]
where $C(a,b)$ is the cosine similarity between the feature-name embeddings, as defined in Section~\ref{sec:identifying_side_effects}. We record the regression's $R^2$. Finally, we report the equally weighted arithmetic mean and minimum--maximum of each statistic across the 16 matrices, expressing $R^2$ as a percentage.

Algorithm~\ref{alg:feature_independence_quantification} summarizes the calculation procedure.

\begin{algorithm}[H]

\caption{Cross-Feature Magnitude, Prevalence, and Similarity}
\label{alg:feature_independence_quantification}
\small
\begin{algorithmic}
\STATE \textbf{Input:} Win-rate matrices $W_{g,d}$, feature-name cosine similarities $C$, and judgments grouped by initiating message.
\STATE \textbf{Output:} Mean and range of four statistics across model--domain matrices.
\STATE $\mathcal{P}\gets\{(a,b):a,b\in\mathcal{F},\ a\ne b\}$; $|\mathcal{P}|=156$.
\FOR{each model--domain matrix $(g,d)$}
    \FOR{each $(a,b)\in\mathcal{P}$}
        \STATE $D_{g,d}(a,b)\gets100|W_{g,d}(a,b)-0.5|$.
        \STATE For each initiating message $i$, average its repeated judgments to obtain $\bar{y}_i$.
        \STATE $p_{g,d}(a,b)\gets$ one-sample $t$-test of $\{\bar{y}_i\}_i$ against $0.5$.

    \ENDFOR
    \STATE Apply Holm correction to the 156 $p$-values within this matrix, yielding $q_{g,d}(a,b)$.
    \STATE $M_{g,d}\gets\frac{1}{156}\sum_{(a,b)\in\mathcal{P}}D_{g,d}(a,b)$.
    \STATE $P_{10,g,d}\gets\frac{100}{156}\sum_{(a,b)\in\mathcal{P}}\mathbf{1}[D_{g,d}(a,b)\geq10]$.
    \STATE $\mathcal{S}_{g,d}\gets\{(a,b)\in\mathcal{P}:D_{g,d}(a,b)\geq10,$
    \STATE \hspace{1em}$q_{g,d}(a,b)<0.05\}$.
    \STATE $P_{10,\mathrm{sig},g,d}\gets100|\mathcal{S}_{g,d}|/156$.
    \STATE Fit OLS with an intercept: $W_{g,d}(a,b)=\alpha_{g,d}+\beta_{g,d}C(a,b)+\epsilon_{g,d}(a,b)$ over $\mathcal{P}$; record $R^2_{g,d}$.

\ENDFOR
\STATE For each statistic, report its arithmetic mean and minimum--maximum range across the 16 matrices. Express $R^2$ as a percentage.
\end{algorithmic}
\end{algorithm}

\paragraph{Interpretation.}
These statistics describe the strength and prevalence of cross-feature shifts, rather than the prevalence of pairs that satisfy the SES gates. The average $R^2$ of $12.9\%$ indicates that feature-name similarity explains a limited share of the observed win-rate variation under the fitted regression. The residual is unexplained variation under this model, not a direct estimate of ``true'' entanglement; it may also reflect nonlinear relationships, other factors, and evaluation noise. The message-clustered, Holm-corrected analysis here is separate from the two-sided binomial gate used in SES.
\par

\endgroup

\section{Details of Side Effect Identification}
\label{appendix:side_effect_identification_details}

Building on the effect matrices above, this section gives the full Side Effect Score derivation and the multiple-comparison sensitivity analysis for Section~\ref{sec:identifying_side_effects}.

\subsection{Side Effect Score and Selection Procedure}

\label{appendix:ses_algorithm}

This appendix gives the full derivation of the Side Effect Score (SES) used in Section~\ref{sec:identifying_side_effects} to rank candidate (\mainFeature{}, \sideFeature{}) pairs per model per domain.

\paragraph{Inputs.}

For each model $g$ and domain $d \in \{\text{Daily}, \text{Task}\}$, SES is computed on the $13 \times 13$ grid indexed by an ordered pair $(a, b)$, where $a$ is the prompted \mainFeature{} and $b$ is the measured \evaluationFeature{}. For each entry in a heatmap in Figure~\ref{fig:combined_heatmap} we have:
\begin{itemize}[noitemsep]
  \item $W(a, b) \in [0, 1]$ --- the pairwise win rate of the styled response (system prompt requesting $a$, length-controlled per Section~\ref{sec:generation_pipeline}) over the neutral baseline, as judged by \gptOssOneTwentyB{} on the $b$-axis.
  \item $p_W(a, b)$ --- the two-sided binomial-test $p$-value for whether $W(a, b)$ deviates significantly from $0.5$.
\end{itemize}
We additionally precompute a fixed $13 \times 13$ semantic-similarity matrix:
\begin{itemize}[noitemsep]
  \item $C(a, b) = \cos(\mathbf{e}_{a}, \mathbf{e}_{b})$ --- cosine similarity of the OpenAI \texttt{text-embedding-3-large} \cite{openai_text_embedding_3_large} embeddings of the feature names. $C$ does not depend on $g$ or $d$; it captures only the prior linguistic relatedness between two style adjectives.
\end{itemize}

\paragraph{MAD-standardised effect sizes.}

We standardise both $W$ and $C$ using robust statistics so a single anomalous pair does not dominate the scale. For win rate, we centre at the no-preference value ($0.5$):
\[
N_{\textsc{wr}}(a, b) = \frac{W(a, b) - 0.5}{1.4826 \cdot \operatorname*{MAD}_{a' \ne b'} W(a', b')}.
\]
For cosine similarity, we mean-centre because cosine values cluster well above $0$:
\[
Z_{\textsc{cs}}(a, b) = \frac{C(a, b) - \overline{C}}{1.4826 \cdot \operatorname*{MAD}_{a' \ne b'} C(a', b')},
\]
where $\overline{C}$ is the off-diagonal mean of $C$ and $\mathrm{MAD}$ denotes the median absolute deviation taken over the off-diagonal entries of the $13 \times 13$ matrix. The factor $1.4826$ converts MAD to a normal-equivalent standard deviation, so $N_{\textsc{wr}}$ and $Z_{\textsc{cs}}$ are comparable across rows and columns even though their underlying scales differ. The diagonal ratings, where $a=b$, are excluded from the MAD and mean estimates because $W(a, a)$ is a self-comparison and is not informative about side effects.

\paragraph{SES formula.}

The Side Effect Score combines the two standardised quantities and two indicator gates:
\begin{equation*}
\begin{aligned}
\mathrm{SES}(a, b) \;=\;& \bigl| N_{\textsc{wr}}(a, b) - Z_{\textsc{cs}}(a, b) \bigr| \\
\;\cdot\;& \min\bigl(|N_{\textsc{wr}}(a, b)|,\ |Z_{\textsc{cs}}(a, b)|\bigr)\\
\;\cdot\;& \mathbf{1}[\,N_{\textsc{wr}}(a, b) \cdot Z_{\textsc{cs}}(a, b) < 0\,] \\
\;\cdot\;& \mathbf{1}[\,p_W(a, b) \le 0.05\,].
\end{aligned}
\end{equation*}
Each factor encodes a specific desideratum:
\begin{enumerate}[label=(\roman*),noitemsep]
  \item $\bigl|N_{\textsc{wr}}(a, b) - Z_{\textsc{cs}}(a, b)\bigr|$ --- \emph{disagreement} between the empirical effect and the prior expectation. Large values mean the win rate departs from what semantic similarity would predict.
  \item $\min(|N_{\textsc{wr}}(a, b)|, |Z_{\textsc{cs}}(a, b)|)$ --- \emph{magnitude floor}: we trust the score only when both standardised quantities are appreciably away from zero. This downweights cases where one of $N_{\textsc{wr}}(a, b)$, $Z_{\textsc{cs}}(a, b)$ is near zero (and the disagreement term is large only because of the other one).
  \item $\mathbf{1}[N_{\textsc{wr}}(a, b) \cdot Z_{\textsc{cs}}(a, b) < 0]$ --- \emph{polarity flip}: the score is non-zero only when the empirical effect is in the \emph{opposite} direction from the centred semantic-similarity prior. This rules out (a) a below-average-similarity pair with a decrease and (b) an above-average-similarity pair with an increase --- both prior-congruent outcomes that are not side effects under our definition. We retain (c) a below-average-similarity pair with an increase and (d) an above-average-similarity pair with a decrease --- both prior-discordant patterns selected by SES.
  \item $\mathbf{1}[p_W(a, b) \le 0.05]$ --- \emph{significance gate}: the win-rate effect must be statistically distinguishable from $0.5$ at the two-sided binomial level.
\end{enumerate}

\paragraph{Polarity and intervention direction.}

The sign of $N_{\textsc{wr}}(a, b)$ determines how the side effect should be intervened against:
\begin{itemize}[noitemsep]
  \item $N_{\textsc{wr}}(a, b) > 0$: prompting $a$ \emph{elevates} $b$ relative to the neutral baseline --- a positive-polarity shift. The mitigation goal is to \emph{suppress} $b$ while preserving $a$; we label these interventions as \textit{suppress\_positive}.
  \item $N_{\textsc{wr}}(a, b) < 0$: prompting $a$ \emph{suppresses} $b$ --- a negative-polarity shift. The mitigation goal is to \emph{enhance} $b$ while preserving $a$; we label these interventions \textit{enhance\_negative}.
\end{itemize}
The polarity label feeds the prompt-intervention template choice (Section~\ref{sec:mitigation}): \promptText{"Please be \{$a$\} but not \{$b$\}"} for \textit{suppress\_positive}, and \promptText{"Please be \{$a$\} but \{$b$\}"} for \textit{enhance\_negative}.

\paragraph{Per-(model, domain) top-1 pair selection.}

For each model in each domain, we rank the $13 \times 12 = 156$ off-diagonal pairs by $\mathrm{SES}(a, b)$ and pick the top-1 as the model's intervention target in that domain. We additionally apply a domain-appropriate filter on the Main Feature so the chosen pair matches the kind of styling a user would realistically request in that context:
\begin{itemize}[noitemsep]
  \item \textbf{Task domain}: the Main Feature $a$ is restricted to utility-oriented features (\textit{efficient, impartial, expert, professional, informative, helpful}). Task-oriented prompts naturally favour these.
  \item \textbf{Daily domain}: the Main Feature $a$ is restricted to interpersonal features (\textit{outgoing, friendly, curious, polite, engaging, empathetic, caring}). Casual/open-domain conversations naturally favour these.
\end{itemize}
This partition is the same utility / interpersonal split discussed in Section~\ref{sec:identifying_side_effects} when describing the heatmap clusters.

\paragraph{Outputs and downstream use.}

For each model in each domain, the top-1 pair under SES becomes the intervention target for Section~\ref{sec:mitigation}; the full set of 16 chosen pairs is summarised in Table~\ref{tab:selected_side_effects}, and the per-method mitigation win rates against those pairs are reported in Appendix~\ref{appendix:mitigation_results}.

\subsection{Multiple-Comparison Robustness}

\paragraph{Multiple-comparison sensitivity.}
\label{appendix:bonferroni_correction}

Each generation-model/domain matrix contains $13\times12=156$ ordered
off-diagonal hypotheses; $a\!\to\!b$ and $b\!\to\!a$ are distinct.  We correct
each of the $8\times2=16$ matrices separately, rather than treating all
$2{,}496$ tests as one family.  For each pair, decisive judgments are encoded
as Styled${}=1$ and Neutral${}=0$, averaged over five generations within each
of $100$ initiating messages, and tested against $0.5$ with a two-sided
one-sample test.  This clustering avoids treating generations sharing a
message and matched Neutral response as independent.

For each 156-test family,
\[
p_{\mathrm{clust}}^{\mathrm{Bonf}}(a,b)=\min\!\left(156\,p_{\mathrm{clust}}(a,b),1\right).
\]
Thus, a pair passes at family-wise $\alpha_{\mathrm{FWER}}=0.05$ only if
\[
p_{\mathrm{clust}}(a,b) \leq \frac{0.05}{156}=0.0003205.
\]
We also report Holm's step-down correction.  Both methods control the
within-matrix family-wise error rate under arbitrary dependence, although
Bonferroni can be conservative.  All other SES components are held fixed, so
correction changes only whether a pair passes the significance gate.

Table~\ref{tab:bonferroni_sensitivity} summarizes the correction results.
Across the 16 matrices, the raw clustered, Holm, and Bonferroni gates retain
$1{,}798$, $1{,}400$, and $1{,}347$ tests, respectively; after the remaining
SES gates, $611$, $445$, and $423$ pairs remain non-zero
under the three significance gates, respectively. Bonferroni retains 14 of 16 selected
targets, with the same top-ranked eligible target in those matrices.  Its SES
vectors have mean Spearman correlation $0.812$ with the raw-clustered vectors
and mean top-five overlap $3.688/5$.

\begin{table}[t]
\centering
\small
\setlength{\tabcolsep}{4.0pt}
\resizebox{\columnwidth}{!}{%
\begin{tabular}{lrrr}
\toprule
Significance gate & Significant tests & Non-zero SES & Targets retained \\
\midrule
Raw clustered & $1{,}798$ & $611$ & $15/16$ \\
Holm & $1{,}400$ & $445$ & $14/16$ \\
Bonferroni & $1{,}347$ & $423$ & $14/16$ \\
\bottomrule
\end{tabular}%
}
\caption{SES sensitivity to within-matrix correction. ``Non-zero SES'' also
requires the magnitude and polarity gates; target retention is measured over
the 16 mitigation targets.}
\label{tab:bonferroni_sensitivity}
\end{table}

The two exceptions differ.  \gptFiveMiniAbbr{}/Task \textit{Professional
$\to$ More Empathetic} fails even the raw clustered test ($p_{\mathrm{clust}}(a,b)=0.059$), owing to
within-message clustering rather than Bonferroni.  \llamaThreeOneAbbr{}/Task
\textit{Professional $\to$ Less Friendly} passes raw clustering ($p_{\mathrm{clust}}(a,b)=0.0025$)
but not Bonferroni ($p_{\mathrm{clust}}^{\mathrm{Bonf}}(a,b)=0.392$), so correction directly removes
this target.

Survival means evidence against a $50\%$ win rate while controlling, at 5\%,
the chance of any false rejection among the 156 tests in that matrix.  It does
not imply a large, practically important, unexpected, or human-validated
effect.  Conversely, non-survival is insufficient evidence under a
conservative family-wise criterion, not evidence for the null.  Retention of
$14/16$ targets therefore supports the main selection pattern subject to the
two exceptions above.

\section{Details of Side Effect Mitigation}
\label{appendix:side_effect_mitigation_details}

Using the side effect pairs selected in the preceding section, this section details the steering implementation, training setup, and full results for Section~\ref{sec:mitigation}.

\subsection{Steering Procedure}

\label{appendix:steering_algo}

Algorithms~\ref{alg:caa_extraction}, \ref{alg:caa_baking}, and \ref{alg:caa_intervention} detail the three phases of steering intervention: vector extraction, layer selection with checkpoint baking, and inference-time generation.

\paragraph{Phase 1: Steering vector extraction.}

Algorithm~\ref{alg:caa_extraction} constructs symmetric contrastive prompt pairs and computes the mean activation difference at a fixed token position. The positive and baseline prompt templates are identical except for the style label, so any activation difference at the same token position is attributable to the style instruction. The training first-message is used only as a grouping and sampling key; it is not inserted into either extraction prompt.

\begin{algorithm}[H]
\caption{Steering Vector Extraction}
\label{alg:caa_extraction}
\small
\begin{algorithmic}

\STATE \textbf{Input:} Training set $\mathcal{S}_{\text{train}}$ of $(t_i, q_i)$ topic / first-message pairs, target model $g$, candidate layers $\mathcal{L}$, style features $\mathcal{F}$.
\STATE \textbf{Output:} Raw steering vectors $\{\mathbf{v}_f^{(\ell)}\}$ for all $f \in \mathcal{F}$, $\ell \in \mathcal{L}$.

\STATE \textbf{Step 1: Symmetric Contrastive Prompt Construction}
\FOR{each feature $f \in \mathcal{F}$}
    \STATE $\mathcal{C}_f \gets \emptyset$
    \FOR{each $(t_i, q_i) \in \mathcal{S}_{\text{train}}$}
        \STATE $\pi_i^{+} \gets$ ``Topic: $t_i$ \textbackslash n Required Style: $\text{UPPER}(f)$ \textbackslash n Response:'';
        \STATE $\pi_i^{-} \gets$ ``Topic: $t_i$ \textbackslash n Required Style: DEFAULT \textbackslash n Response:'';
        \STATE $\mathcal{C}_f \gets \mathcal{C}_f \cup \{(\pi_i^{+}, \pi_i^{-})\}$;
    \ENDFOR
\ENDFOR

\STATE \textbf{Step 2: Activation Capture and Vector Computation}
\FOR{each feature $f \in \mathcal{F}$}
    \FOR{each $(\pi_i^{+}, \pi_i^{-}) \in \mathcal{C}_f$}
        \FOR{each layer $\ell \in \mathcal{L}$}
            \STATE $\mathbf{h}_{i,+}^{(\ell)} \gets g.\text{get\_activation}(\ell, \pi_i^{+}, \text{pos}=-1)$;
            \STATE $\mathbf{h}_{i,-}^{(\ell)} \gets g.\text{get\_activation}(\ell, \pi_i^{-}, \text{pos}=-1)$;
        \ENDFOR
    \ENDFOR
    \FOR{each layer $\ell \in \mathcal{L}$}
        \STATE $\mathbf{v}_f^{(\ell)} \gets \frac{1}{|\mathcal{C}_f|} \sum_i (\mathbf{h}_{i,+}^{(\ell)} - \mathbf{h}_{i,-}^{(\ell)})$;
    \ENDFOR
\ENDFOR

\end{algorithmic}
\end{algorithm}

\paragraph{Phase 2: Layer selection and model baking.}

Once raw vectors are computed, we identify the best injection layer using the validation split and permanently bake the chosen vector into the model weights (Algorithm~\ref{alg:caa_baking}).

\begin{algorithm}[H]
\caption{Validation and Checkpoint Baking}
\label{alg:caa_baking}
\small
\begin{algorithmic}

\STATE \textbf{Input:} Raw vectors $\{\mathbf{v}_f^{(\ell)}\}$, validation set $\mathcal{S}_{\text{val}}$, target model $g$, judge model $M_{\text{judge}}$.
\STATE \textbf{Output:} Best layers $\{\ell^*_f\}$, baked checkpoints $\{g_f^{\text{steered}}\}$.

\STATE \textbf{Step 1: Best Layer Selection}
\FOR{each feature $f \in \mathcal{F}$}
    \FOR{each layer $\ell \in \mathcal{L}$}
        \STATE $g^{(\ell)}_f \gets g$ with $\mathbf{v}_f^{(\ell)}$ added to layer $\ell$ activations at the last token position during the forward pass;
        \FOR{each $(t_i, q_i, y_i^{\text{neutral}}) \in \mathcal{S}_{\text{val}}$}
            \FOR{repetition $r = 1$ to $5$}
                \STATE $y_{i,r}^{\text{steered}} \gets g^{(\ell)}_f.\text{generate}(q_i)$;
                \STATE $w_{i,r} \gets M_{\text{judge}}.\text{compare}(f, y_{i,r}^{\text{steered}}, y_i^{\text{neutral}})$;
            \ENDFOR
        \ENDFOR
        \STATE $W_{\mathrm{val}}(f,\ell) \gets \frac{1}{5|\mathcal{S}_{\text{val}}|} \sum_i \sum_{r=1}^{5} \mathbf{1}[w_{i,r} = \text{steered}]$;
    \ENDFOR
    \STATE $\ell^*_f \gets \arg\max_{\ell \in \mathcal{L}} W_{\mathrm{val}}(f,\ell)$;
\ENDFOR

\STATE \textbf{Step 2: Checkpoint Baking}
\FOR{each feature $f \in \mathcal{F}$}
    \STATE $\theta \gets \text{load\_checkpoint}(g)$;
    \STATE $\theta[\texttt{layers.}\ell^*_f\texttt{.mlp.down\_proj.bias}] \gets \mathbf{v}_f^{(\ell^*_f)}$;
    \STATE \quad // \texttt{down\_proj} has no bias by default; setting it to $\mathbf{v}_f^{(\ell^*_f)}$ is equivalent to adding the vector at the layer's output during the forward pass.
    \STATE Set \texttt{mlp\_bias=True} in the model config; save as $g_f^{\text{steered}}$;
\ENDFOR

\end{algorithmic}
\end{algorithm}

\paragraph{Phase 3: Test-set intervention.}

In Algorithm~\ref{alg:caa_intervention} we load the baked checkpoint corresponding to the targeted \emph{Side Feature} $b$ and generate test responses while the prompt still asks for the \emph{Main Feature} $a$. The side effect mitigation thus combines a styled-prompt signal (from $a$) with a baked-in activation shift toward $b$ (or away from it, by polarity).

\begin{algorithm}[H]
\caption{Steering on Test Set}
\label{alg:caa_intervention}
\small
\begin{algorithmic}

\STATE \textbf{Input:} Baked checkpoints $\{g_f^{\text{steered}}\}$, test set $\mathcal{S}_{\text{test}}$, per-(model, domain) targeted tuple $(a, b, \sigma)$.
\STATE \textbf{Output:} Steered generations $\{y_{i,r}^{\text{steered}}\}$.

\STATE $g^{\text{steered}} \gets g_b^{\text{steered}}$ (with polarity sign $\sigma$ applied to $\mathbf{v}_b^{(\ell^*_b)}$ before baking);
\FOR{each $(t_i, q_i) \in \mathcal{S}_{\text{test}}$}
    \FOR{repetition $r = 1$ to $5$}
        \STATE $\pi_i^{\mathrm{sys}} \gets$ ``You are having a conversation about $t_i$. Please be $a$ in your responses. Please answer in around $L_{q_i,g}$ words. Respond directly without showing your reasoning.'';
        \STATE $y_{i,r}^{\text{steered}} \gets g^{\text{steered}}.\text{generate}(\pi_i^{\mathrm{sys}}, q_i)$;
        \STATE \quad // Prompt requests $a$; the activation shift toward or away from $b$ is baked into the \texttt{down\_proj} bias of layer $\ell^*_b$.
    \ENDFOR
\ENDFOR

\end{algorithmic}
\end{algorithm}

\paragraph{Notation and definitions.}

\noindent \textbf{Target models:} the 6 open-weight models for which steering is applied --- \llamaThreeOneAbbr{}, \qwenNineBAbbr{}, \qwenTwentySevenBAbbr{}, \gemmaThirtyOneBAbbr{}, \gemmaEfourBAbbr{}, and \gptOssTwentyBAbbr{}. The two API-served models (\geminiFlashLiteAbbr{}, \gptFiveMiniAbbr{}) do not expose internal activations and are excluded from steering. The judge is \gptOssOneTwentyB{}.

\noindent \textbf{Steering vector $\mathbf{v}_f^{(\ell)}$:} a vector in $\mathbb{R}^{d_{\text{model}}}$ representing the direction in activation space that increases feature $f$ at layer $\ell$. Computed as the mean of $480$ per-pair differences $\mathbf{h}_{i,+}^{(\ell)} - \mathbf{h}_{i,-}^{(\ell)}$ across the training split of the (model, domain) pairs.

\noindent \textbf{Contrastive prompt pair:} two prompts that differ only in the style label, formatted as ``\texttt{Topic: \{topic\}\textbackslash n Required Style: \{label\}\textbackslash n Response:}'' where \texttt{label} is the uppercase feature name on the positive branch and \texttt{DEFAULT} on the baseline branch. The first-message is a grouping key and is not appended to the prompts. No chat template is applied at extraction time; the prompts are raw completion-style scaffolds.

\noindent \textbf{Activation extraction:} activations are captured at the final token position ($\text{pos}=-1$, the \texttt{:} after \texttt{Response}) of each prompt at every candidate layer.

\noindent \textbf{Candidate layers $\mathcal{L}$:} auto-selected as the middle $60\%$ of the model's transformer layers, evenly spaced into $7$ candidates. For \llamaThreeOneAbbr{} (32 layers) this yields $\mathcal{L} = \{6, 9, 12, 15, 18, 21, 24\}$; the exact indices vary by architecture depth.

\noindent \textbf{Best-layer selection:} for each feature $f$, $\ell^*_f = \arg\max_{\ell \in \mathcal{L}} W_{\mathrm{val}}(f,\ell)$, where $W_{\mathrm{val}}(f,\ell)$ is the validation-set win rate of the steered model against the \textit{neutral} response on the side-feature axis. The styled ground-truth response is \emph{not} used in selection.

\noindent \textbf{Checkpoint baking:} the steering vector $\mathbf{v}_f^{(\ell^*_f)}$ is written to the bias slot of the layer's MLP \texttt{down\_proj} ($\theta[\texttt{layers.}\ell^*_f\texttt{.mlp.down\_proj.bias}] \gets \mathbf{v}_f^{(\ell^*_f)}$). Since the base model has no bias on this projection, setting the slot is equivalent to adding the vector to the layer's output at inference, with no runtime hook required.

\noindent \textbf{Targeted tuple $(a, b, \sigma)$:} per-(model, domain) pair, the top-SES pair (Section~\ref{sec:identifying_side_effects}). The polarity sign $\sigma$ is applied to the side-feature vector before baking: positive polarity (main over-elevates side $\Rightarrow$ we want side down) uses $-\mathbf{v}_b^{(\ell^*_b)}$; negative polarity (main suppresses side $\Rightarrow$ we want side up) uses $+\mathbf{v}_b^{(\ell^*_b)}$.

\subsection{Training, Evaluation, and Full Results}

\label{appendix:training_setup}

We use the following training procedure for the DPO and SFT
interventions in Section~\ref{sec:mitigation}. All training was run on the
intervention split of 4,800 first-message pairs per model in each domain
(480 training first-messages $\times$ 10 repetitions; the matching 160 val
and 160 test first-messages are held out for the figures and tables in the
main paper).

\paragraph{Frameworks.}

We use two open-source training stacks, selected per model:

\begin{itemize}[noitemsep]
  \item \textbf{LLaMA-Factory} (with DeepSpeed ZeRO-3) for the full-model
        fine-tuning of \texttt{Llama-3.1-8B} and \texttt{Qwen3.5-9B} on
        $8\times$A6000. We patched only \code{dataset\_info.json} to
        register the JSON files for each model in each domain.
  \  \item \textbf{Unsloth} on a single B200 for the
        larger / specially-quantized models: \texttt{Qwen3.5-27B},
        \texttt{Gemma-4-31B}, and \texttt{gpt-oss-20b} (QLoRA-4bit), and
        \texttt{Gemma-4-E4B} (bf16 full-FT). We load each model from
        Unsloth's pre-processed checkpoints
        (\code{unsloth/gemma-4-31B-it},
        \code{unsloth/gemma-4-E4B-it},
        \href{https://huggingface.co/unsloth/Qwen3.5-27B-Instruct-bnb-4bit}{\code{unsloth/}\allowbreak\code{Qwen3.5-27B-Instruct-}\allowbreak\code{bnb-4bit}}, and
        \code{unsloth/gpt-oss-20b}) and apply the \code{gemma-4}
        chat template via \code{get\_chat\_template} for the Gemma-4
        models and the model's native template for the rest. LoRA
        adapters (rank $16$, $\alpha_{\mathrm{LoRA}}{=}32$) are attached to
        \texttt{q\_proj, k\_proj, v\_proj, o\_proj, gate\_proj,
        up\_proj, down\_proj}; for \texttt{gpt-oss-20b} we list these
        explicitly rather than relying on the \code{"all-linear"}
        shorthand so the MoE expert projections are included.
\end{itemize}

\paragraph{Training data construction.}

For each model and domain we construct a single JSON dataset:

\begin{itemize}[noitemsep]
  \item \textbf{DPO}: \texttt{chosen} = the prompting-intervention response
        (\textit{Please be \{main\} but not \{side\}} for
        suppress\_positive cases, \textit{Please be \{main\} but \{side\}}
        for enhance\_negative cases); \texttt{rejected} = the styled-only
        response (\textit{Please be \{main\}}). Same
        \texttt{first\_message}, same model, same per-message length budget
        $L_{q_i,g}$ (Section~\ref{sec:generation_pipeline}). 4{,}800 preference pairs
        per model and domain.
  \item \textbf{SFT}: single-target supervised pairs; target = intervened
        response from the prompting intervention. 4{,}800 examples per model
        and domain.
\end{itemize}

The styled system prompt is embedded inside every training example as a
\code{system} message so that test-time inference (which queries the trained
checkpoint with the same styled prompt) sees the same context distribution
the model was trained on. Without this we observed near-zero transfer: the
trained checkpoint would silently fall back to the model's default system
template at inference time and produce the original styled response rather
than the intervened one.

\paragraph{Hyperparameters.}

Shared across every reported models: \textbf{1 epoch}, AdamW with a cosine
schedule and warmup ratio $0.1$, bf16 mixed precision, and (for DPO)
$\beta = 0.1$ with the sigmoid loss. We pick epoch=1 because we are limited by compute constraint. The learning rate and fine-tuning
class differ by model and method as listed in
Table~\ref{tab:training_per_model}; the two stacks use different effective
batch sizes (LLaMA-Factory: $\approx$64 across $8$ GPUs; Unsloth:
per-device batch $1\times$ gradient accumulation $8$ on a single B200).
QLoRA use LoRA rank $16$, $\alpha_{\mathrm{LoRA}}{=}32$. We report the full-model
checkpoints for \texttt{Llama-3.1-8B} and \texttt{Qwen3.5-9B} (directly
comparable to each other); \texttt{Qwen3.5-27B}, \texttt{Gemma-4-31B}, and
\texttt{gpt-oss-20b} are adapter/QLoRA-only because full-model training of
these does not fit our hardware budget.

\begin{table}[t]
\centering
\resizebox{\columnwidth}{!}{%

\begin{tabular}{@{} l l l c c @{}}

\toprule

\textbf{Model} & \textbf{Stack} & \textbf{FT class} & \textbf{SFT lr} & \textbf{DPO lr} \\

\midrule

Llama-3.1-8B  & LLaMA-Factory & full-FT     & $2{\times}10^{-5}$ & $5{\times}10^{-6}$ \\

Qwen3.5-9B    & LLaMA-Factory & full-FT     & $2{\times}10^{-5}$ & $5{\times}10^{-6}$ \\

Qwen3.5-27B   & Unsloth       & QLoRA-4bit  & $1{\times}10^{-4}$ & $5{\times}10^{-6}$ \\

Gemma-4-31B   & Unsloth       & QLoRA-4bit  & $1{\times}10^{-4}$ & $5{\times}10^{-6}$ \\

Gemma-4-E4B   & Unsloth       & full-FT$^\dagger$ & $2{\times}10^{-6}$ & $2{\times}10^{-6}$ \\

gpt-oss-20b   & Unsloth       & QLoRA-4bit  & $1{\times}10^{-4}$ & $5{\times}10^{-6}$ \\

\bottomrule

\end{tabular}%

}

\caption{Per-model training configuration behind the reported numbers. All runs use 1 epoch, cosine schedule, warmup $0.1$, bf16, and sequence length $4096$. LLaMA-Factory configurations are full-model fine-tuned with DeepSpeed ZeRO-3 on $8\times$A6000; Unsloth configurations (LoRA rank $16$, $\alpha_{\mathrm{LoRA}}{=}32$ for QLoRA) run on a single B200. $^\dagger$Gemma-4-E4B is full-FT (with vision-tower freeze + fp32 mixed-precision) for 3 of its 4 configurations; the task/SFT configuration is unstable in full-FT (loss diverges to NaN regardless of LR, gradient clipping, or precision) and falls back to a LoRA adapter (SFT lr $1{\times}10^{-4}$). gpt-oss-20b is natively MXFP4 and uses explicit attention+MoE LoRA targets.}
\label{tab:training_per_model}
\end{table}

\paragraph{Compute.}

Each run finishes in 1--3 hours on $8 \times$ NVIDIA A6000 (48~GB,
LLaMA-Factory full-FT) or a single NVIDIA B200 (180~GB, Unsloth). The full
24-run sweep (6 trainable models $\times$ 2 domains $\times$ 2 methods,
excluding the 2 API-served models, which we run with prompt intervention
only) consumed on the order of a few hundred GPU-hours across a mixed
A6000 / B200 cluster.

\paragraph{Evaluation.}

After each training run, we run vLLM inference on the 160-first-message
test split (5 repetitions per message, \texttt{temperature}\(=0.7\) for local
models) and rate each \texttt{(intervened, baseline)} pair with the
\texttt{GPT-OSS-120B} judge via the same prompt as in Appendix~\ref{appendix:prompt_examples}. The resulting
win rates power Figure~\ref{fig:merged_comparison}.

\paragraph{Mitigation results: full win rates.}
\label{appendix:mitigation_results}

Table~\ref{tab:MSE_results} reports the full win-rate numbers behind
Figure~\ref{fig:merged_comparison}.

\begin{table*}[!t]
    \centering
    \scriptsize
    \renewcommand{\arraystretch}{0.95}
    \begin{tabular}{@{}cclcrrrrr@{}}
        \toprule
        \textbf{Domain} & \textbf{Model} & \textbf{Side Effect (Main\,$\to$\,Side)} & \textbf{Feat.} & \textbf{BEFORE} & \textbf{Prompt} & \textbf{Steer} & \textbf{DPO} & \textbf{SFT} \\
        \midrule
        \multirow{16}{*}{Daily}
          & \multirow{2}{*}{\texttt{Gemini3.1-FL}} & \multirow{2}{*}{friendly\,$\to$\,professional\,$\downarrow$}
              & Main & 0.70 & 0.83 & --- & --- & --- \\
          &                                       &
              & Side & 0.36 & 0.81 & --- & --- & --- \\
        \cmidrule(l){2-9}
          & \multirow{2}{*}{\texttt{Gemma4-31B}}   & \multirow{2}{*}{friendly\,$\to$\,professional\,$\downarrow$}
              & Main & 0.97 & 0.73 & 0.59 & 0.77 & 0.73 \\
          &                                       &
              & Side & 0.16 & 0.61 & 0.31 & 0.17 & 0.57 \\
        \cmidrule(l){2-9}
          & \multirow{2}{*}{\texttt{Gemma4-E4B}}   & \multirow{2}{*}{friendly\,$\to$\,professional\,$\downarrow$}
              & Main & 0.84 & 0.62 & 0.57 & 0.53 & 0.51 \\
          &                                       &
              & Side & 0.23 & 0.59 & 0.49 & 0.39 & 0.51 \\
        \cmidrule(l){2-9}
          & \multirow{2}{*}{\texttt{GPT-OSS-20B}}  & \multirow{2}{*}{friendly\,$\to$\,professional\,$\downarrow$}
              & Main & 0.87 & 0.64 & 0.64 & 0.74 & 0.58 \\
          &                                       &
              & Side & 0.22 & 0.64 & 0.44 & 0.37 & 0.52 \\
        \cmidrule(l){2-9}
          & \multirow{2}{*}{\texttt{GPT-5-mini}}   & \multirow{2}{*}{empathetic\,$\to$\,outgoing\,$\uparrow$}
              & Main & 0.84 & 0.76 & --- & --- & --- \\
          &                                       &
              & Side & 0.67 & 0.52 & --- & --- & --- \\
        \cmidrule(l){2-9}
          & \multirow{2}{*}{\texttt{Llama3.1-8B}}  & \multirow{2}{*}{curious\,$\to$\,informative\,$\downarrow$}
              & Main & 0.87 & 0.67 & 0.48 & 0.47 & 0.49 \\
          &                                       &
              & Side & 0.34 & 0.57 & 0.58 & 0.85 & 0.48 \\
        \cmidrule(l){2-9}
          & \multirow{2}{*}{\texttt{Qwen3.5-9B}}   & \multirow{2}{*}{friendly\,$\to$\,professional\,$\downarrow$}
              & Main & 0.94 & 0.77 & 0.78 & 0.43 & 0.70 \\
          &                                       &
              & Side & 0.15 & 0.58 & 0.38 & 0.89 & 0.62 \\
        \cmidrule(l){2-9}
          & \multirow{2}{*}{\texttt{Qwen3.5-27B}}  & \multirow{2}{*}{friendly\,$\to$\,professional\,$\downarrow$}
              & Main & 0.95 & 0.76 & 0.73 & 0.50 & 0.69 \\
          &                                       &
              & Side & 0.09 & 0.60 & 0.41 & 0.41 & 0.53 \\
        \midrule
        \multirow{16}{*}{Task}
          & \multirow{2}{*}{\texttt{Gemini3.1-FL}} & \multirow{2}{*}{impartial\,$\to$\,professional\,$\uparrow$}
              & Main & 0.61 & 0.51 & --- & --- & --- \\
          &                                       &
              & Side & 0.55 & 0.57 & --- & --- & --- \\
        \cmidrule(l){2-9}
          & \multirow{2}{*}{\texttt{Gemma4-31B}}   & \multirow{2}{*}{impartial\,$\to$\,professional\,$\uparrow$}
              & Main & 0.75 & 0.39 & 0.59 & 0.44 & 0.29 \\
          &                                       &
              & Side & 0.55 & 0.20 & 0.53 & 0.15 & 0.19 \\
        \cmidrule(l){2-9}
          & \multirow{2}{*}{\texttt{Gemma4-E4B}}   & \multirow{2}{*}{expert\,$\to$\,efficient\,$\downarrow$}
              & Main & 0.60 & 0.61 & 0.40 & 0.71 & 0.60 \\
          &                                       &
              & Side & 0.45 & 0.50 & 0.51 & 0.46 & 0.48 \\
        \cmidrule(l){2-9}
          & \multirow{2}{*}{\texttt{GPT-OSS-20B}}  & \multirow{2}{*}{professional\,$\to$\,friendly\,$\downarrow$}
              & Main & 0.60 & 0.52 & 0.46 & 0.60 & 0.58 \\
          &                                       &
              & Side & 0.47 & 0.66 & 0.55 & 0.49 & 0.55 \\
        \cmidrule(l){2-9}
          & \multirow{2}{*}{\texttt{GPT-5-mini}}   & \multirow{2}{*}{professional\,$\to$\,empathetic\,$\uparrow$}
              & Main & 0.57 & 0.52 & --- & --- & --- \\
          &                                       &
              & Side & 0.46 & 0.37 & --- & --- & --- \\
        \cmidrule(l){2-9}
          & \multirow{2}{*}{\texttt{Llama3.1-8B}}  & \multirow{2}{*}{professional\,$\to$\,friendly\,$\downarrow$}
              & Main & 0.62 & 0.58 & 0.40 & 0.35 & 0.51 \\
          &                                       &
              & Side & 0.45 & 0.59 & 0.69 & 0.86 & 0.51 \\
        \cmidrule(l){2-9}
          & \multirow{2}{*}{\texttt{Qwen3.5-9B}}   & \multirow{2}{*}{professional\,$\to$\,friendly\,$\downarrow$}
              & Main & 0.66 & 0.49 & 0.48 & 0.31 & 0.53 \\
          &                                       &
              & Side & 0.40 & 0.73 & 0.66 & 0.96 & 0.68 \\
        \cmidrule(l){2-9}
          & \multirow{2}{*}{\texttt{Qwen3.5-27B}}  & \multirow{2}{*}{professional\,$\to$\,friendly\,$\downarrow$}
              & Main & 0.71 & 0.51 & 0.62 & 0.31 & 0.51 \\
          &                                       &
              & Side & 0.37 & 0.77 & 0.59 & 0.64 & 0.71 \\
        \bottomrule
    \end{tabular}
    \caption{Side effect mitigation results across all 8 generation models and both domains. Each entry reports the top side effect pair (Main\,$\to$\,Side) identified per (model, domain) by the side effect score, and the win-rate proportion \(W\in[0,1]\) of the model's response over the no-style neutral baseline on both the Main and Side feature axes. \textbf{BEFORE} = styled response with only the main-feature prompt (no mitigation). \textbf{Prompt} / \textbf{Steer} / \textbf{DPO} / \textbf{SFT} = the four mitigation methods. "\,$\downarrow$\," marks a negative-polarity side effect (Main Feature suppresses the Side Feature; successful mitigation raises the Side win rate); "\,$\uparrow$\," marks a positive-polarity side effect (Main Feature elevates the Side Feature; successful mitigation lowers it). "---" indicates the method was not run for that (model, domain) setting: Steering, DPO and SFT were not run for the API-served models (\texttt{Gemini3.1-FL}, \texttt{GPT-5-mini}). Same numbers as Figure~\ref{fig:merged_comparison}.}
    \label{tab:MSE_results}
\end{table*}

For each model--domain pair, it lists the top SES pair and the Main- and
Side-feature win rates before mitigation and after Prompt, Steering, DPO,
and SFT. For the API-served models (\texttt{Gemini3.1-FL} and
\texttt{GPT-5-mini}), only Prompting was evaluated.

\section{Response-Difference Analysis of Side-Effect Mechanisms}

\label{app:response_difference_analysis}

\subsection{Analysis Protocol}
\label{app:response_difference_method}
\paragraph{Data.} We analyze two paired comparisons for existing Gemma-4-E4B generations: styled versus baseline and styled versus prompt-mitigated. Both cover the Friendly$\rightarrow$less-Professional pair in the Daily domain and the Expert$\rightarrow$less-Efficient pair in the Task domain. In the styled--baseline comparison, each setting contains 500 pairs from 100 unique user prompts, with five styled generations per prompt. The same original response, generated without the target-style instruction, is paired with all five styled generations for that prompt, yielding 1,000 pairs across the two settings.

The styled--prompt-mitigated comparison uses the existing natural-order prompt-intervention test records, in which the mitigation instruction jointly requests the Main Feature and the suppressed Side Feature (Friendly followed by Professional, or Expert followed by Efficient). Each setting contains 800 matched pairs from 160 unique user prompts, with five repetitions per prompt, yielding 1,600 pairs across the two settings. Each prompt-mitigated response is paired with the styled response stored in the same intervention record; thus, the domain, target feature pair, user prompt, and repetition are matched. No Gemma responses are regenerated. The mitigation comparison uses disjoint pair identifiers and a separate annotation checkpoint, so running it neither resubmits nor overwrites the completed styled--baseline annotations.

\paragraph{Difference extraction.} GPT-OSS-120B annotates every response pair using the same protocol. At the lexical/style level, a first pass extracts candidate adjectives and visible punctuation or formatting marks from each response, and a second, conservative quality-control pass removes candidates that are not adjectives in context. The original user prompt is omitted from both lexical passes. At the content level, a third pass jointly examines both responses with the original user prompt, aligns semantically equivalent atomic content, and identifies content present in only one response. The annotation model is run at temperature~0 with low reasoning effort.

For a given comparison, let $y^{\mathrm{styled}}_i$ denote the styled response and $y^{\mathrm{cmp}}_i$ its comparator: the baseline response in the first comparison and the prompt-mitigated response in the second.

For each extraction family $k\in\{\text{adjective},\text{punctuation},\text{content}\}$, let $E_k(y)$ denote the items extracted from response $y$. Specifically, $E_{\text{adjective}}(y)$ is the set of quality-controlled adjectives in $y$; $E_{\text{punctuation}}(y)$ is the set of punctuation and formatting markers in $y$; and $E_{\text{content}}(y)$ is the set of atomic content units assigned to $y$ after jointly aligning the paired responses. The notation $E_{\text{content}}(y)$ is shorthand for a pair-conditioned extraction that also depends on the paired response and the original prompt $q_i$, unlike the lexical extraction functions.

For response pair $i$, we define the styled-only and comparator-only sets as

\begin{align}
    D^{+}_{i,k} &= E_k(y^{\mathrm{styled}}_i) \setminus E_k(y^{\mathrm{cmp}}_i), \\
    D^{-}_{i,k} &= E_k(y^{\mathrm{cmp}}_i) \setminus E_k(y^{\mathrm{styled}}_i).
\end{align}

Thus, $D^{+}_{i,k}$ always contains styled-only items; $D^{-}_{i,k}$ contains baseline-only items in the first comparison and prompt-mitigated-only items in the second. For atomic content, the notation refers to unmatched content units produced after semantic alignment rather than literal string-level set subtraction.

\paragraph{Aggregation and association.} We compute all summaries separately for each target setting and comparison; accordingly, $N=500$ for a styled--baseline setting and $N=800$ for a styled--prompt-mitigated setting. We report the prevalence of an item $x$ as the proportion of paired generations whose corresponding difference set contains $x$:

\begin{equation}
    \operatorname{Prev}^{\pm}_k(x) = \frac{1}{N} \sum_{i=1}^{N} \mathbf{1}\!\left[x\in D^{\pm}_{i,k}\right].
\end{equation}

Within each topic, we additionally report the mean sizes of the styled-only and comparator-only sets. These quantities summarize how much distinct material appears on each side; they are not signed treatment-effect estimates.

We also retain the existing pairwise Main and Side Feature verdicts; each results table states explicitly which response's win rate is reported. For the association analysis, let $Y_i=1$ indicate that the judge preferred the comparator on the Side Feature. In the styled--baseline comparison, this means that the baseline was preferred on Professional or Efficient, indicating that the styled response exhibits the selected side effect. In the styled--prompt-mitigated comparison, it means that the prompt-mitigated response was preferred on Professional or Efficient, indicating Side Feature recovery relative to the matched styled response. To identify response differences associated with the corresponding verdict, we compute

\begin{equation}
\begin{aligned}
    \Delta^{\pm}_{\mathrm{assoc},k}(x)
    &= \Pr\!\left(x\in D^{\pm}_{i,k}\mid Y_i=1\right) \\
    &\quad - \Pr\!\left(x\in D^{\pm}_{i,k}\mid Y_i=0\right).
\end{aligned}
\end{equation}

Items supported by at least three paired generations are ranked by $\lvert\Delta^{\pm}_{\mathrm{assoc},k}(x)\rvert$. These prevalence differences are descriptive associations; they are not causal effects or significance tests. Because atomic-content descriptions are canonicalized within, but not across, response pairs, their cross-pair prevalence is wording-sensitive and is interpreted conservatively. Rows that share a user prompt are repeated paired generations, with the pair---rather than the unique prompt---as the unit of these descriptive summaries.

\subsection{Overall Outcomes}

Table~\ref{tab:response_difference_overview} summarizes how the Styled conditions tend to strengthen the Main Feature while introducing the selected Side Feature shift. Table~\ref{tab:response_difference_mitigation_overview} shows the complementary mitigation trade-off: relative to its matched Styled response, the prompt-mitigated response is preferred more often on the Side Feature and less often on the Main Feature.

\begin{table}[t]

    \centering

    \small

    \setlength{\tabcolsep}{3pt}
    \resizebox{\columnwidth}{!}{%
    \begin{tabular}{lrrrr}

        \toprule

        Main / Side Feature & Domain & Pairs & Main Win & Side Effect \\

        \midrule

        Friendly / Professional & Daily & 500 & 95.8\% & 87.0\% \\

        Expert / Efficient & Task & 500 & 68.8\% & 66.4\% \\

        \bottomrule

    \end{tabular}%
    }

    \caption{Overview of the response-difference case studies. ``Main Win'' denotes that the styled response was preferred on the prompted feature. ``Side Effect'' denotes that the baseline was preferred as more Professional or more Efficient.}

    \label{tab:response_difference_overview}

\end{table}

\begin{table}[t]

    \centering

    \small

    \setlength{\tabcolsep}{5pt}
    \resizebox{\columnwidth}{!}{%
    \begin{tabular}{lrr}

        \toprule

        Main / Side Feature & Main Win & Side Win \\

        \midrule

        Friendly / Professional & 18.5\% & 88.1\% \\

        Expert / Efficient & 47.6\% & 55.8\% \\

        \bottomrule

    \end{tabular}%
    }

    \caption{Styled--prompt-mitigated outcomes. The Main Win and Side Win columns report prompt-mitigated preference rates over the exactly matched Styled response. The separate marginal preferences show Side Feature recovery together with lower relative Main Feature strength.}

    \label{tab:response_difference_mitigation_overview}

\end{table}

\subsection{Friendly $\rightarrow$ Less Professional}

\label{app:friendly_professional_analysis}

\paragraph{Lexical and punctuation changes.} Friendly responses exhibit a consistent shift toward expressive language. Common styled-only adjectives include \emph{wonderful}, \emph{great}, \emph{exciting}, \emph{interesting}, and \emph{lovely}. The dominant punctuation change is the exclamation mark, which occurs in the styled-only set of 62.0\% of all Friendly pairs. Styled-only question marks and apostrophes are also common, consistent with more follow-up questions and contracted conversational forms.

More importantly, these punctuation changes are associated with the side-effect verdict. As shown in Table~\ref{tab:friendly_professional_association}, an exclamation mark is much more frequent when the styled response is judged less Professional. Commas, question marks, and apostrophes show smaller but directionally similar associations.

\begin{table}[t]

    \centering

    \small

    \setlength{\tabcolsep}{3pt}

    \resizebox{\columnwidth}{!}{%
    \begin{tabular}{lrrrr}

        \toprule

        Styled-only Item & $n$ & $Y{=}1$ & $Y{=}0$ & $\Delta_{\mathrm{assoc}}$ \\

        \midrule

        \texttt{!} & 310 & 66.9\% & 29.2\% & $+37.7$ pp \\

        \texttt{,} & 96 & 20.7\% & 9.2\% & $+11.5$ pp \\

        \texttt{?} & 106 & 22.3\% & 13.8\% & $+8.5$ pp \\

        \texttt{'} & 110 & 22.8\% & 16.9\% & $+5.8$ pp \\

        \bottomrule

    \end{tabular}%
    }

    \caption{Styled-only punctuation associated with the Friendly$\rightarrow$less-Professional side effect. $Y=1$ indicates that the side effect was observed, and $\Delta_{\mathrm{assoc}}$ is the prevalence difference between $Y=1$ and $Y=0$ pairs. Values are descriptive rather than causal.}

    \label{tab:friendly_professional_association}

\end{table}

\paragraph{Atomic-content changes.} Friendly responses add interactional moves such as expressing concern, advising the user to take care, asking what happened, offering additional help, wishing the user well, or asking a conversational follow-up. Although their wording adapts to the input, these additions share a functional pattern: they extend the social interaction beyond the minimum propositional answer.

\paragraph{Interpretation.} The results indicate that the side effect is primarily a register-level phenomenon. Gemma-4-E4B realizes Friendliness by increasing enthusiasm, emotional engagement, and conversational continuation. These changes are effective signals of Friendliness but can simultaneously make the response appear less restrained or less formal. The mechanism is therefore not the removal of one Professional content element, but the accumulation of expressive and interactional cues associated with an informal conversational register. The side effect appears in every Daily topic, suggesting that this realization pattern is not confined to a single subject area.

\paragraph{Prompt mitigation as selective bundle reweighting.} The added Professional instruction produces the Main/Side preference trade-off shown in Table~\ref{tab:response_difference_mitigation_overview}. Each matched response answers the same user request, and the two aggregate one-sided lists share everyday evaluative adjectives such as \emph{great}, \emph{wonderful}, and \emph{lovely}. The difference lies in concentration: these terms, together with \emph{fun}, are more prominent in the styled-only set, and Professional recovery is associated with \emph{wonderful}, \emph{great}, and \emph{fun} remaining on that side. The content differences provide a parallel functional contrast. Styled-only additions extend the social exchange through anticipation, well wishes, or conversational follow-ups, whereas prompt-mitigated-only additions orient the same request toward direct assistance and concrete task details. Thus, the mitigation condition remains aligned with the same request while selectively attenuating overt warmth. Across the matched comparison, this reweighting coincides with Professional recovery and lower Friendly preference.

\subsection{Expert $\rightarrow$ Less Efficient}

\label{app:expert_efficient_analysis}

\paragraph{Aggregate response differences.} The Expert instruction produces a broad lexical shift. The most frequent styled-only adjectives include \emph{specific}, \emph{complex}, \emph{crucial}, \emph{robust}, \emph{significant}, \emph{deep}, and \emph{critical}. Styled responses also introduce more parenthetical punctuation, hyphens, commas, and em-dashes, reflecting greater internal structuring and qualification.

These aggregate changes describe how the model expresses Expertise through a collection of lexical and formatting cues. Table~\ref{tab:expert_efficient_association} highlights depth-oriented terms associated with the less-Efficient judgment. Together with the broader content additions below, these associations link explicit depth and qualification to the Side Feature shift.

\begin{table}[t]

    \centering

    \small

    \setlength{\tabcolsep}{3pt}

    \begin{tabular}{lrrrr}

        \toprule

        Styled-only Item & $n$ & $Y{=}1$ & $Y{=}0$ & $\Delta_{\mathrm{assoc}}$ \\

        \midrule

        \emph{critical} & 21 & 5.1\% & 2.4\% & $+2.7$ pp \\

        \emph{detailed} & 12 & 3.3\% & 0.6\% & $+2.7$ pp \\

        \emph{nuanced} & 9 & 2.7\% & 0.0\% & $+2.7$ pp \\

        \bottomrule

    \end{tabular}

    \caption{Selected styled-only depth terms positively associated with the Expert$\rightarrow$less-Efficient judgment.}

    \label{tab:expert_efficient_association}

\end{table}

\paragraph{Atomic-content changes.} Expert responses introduce task-specific content such as additional causal mechanisms, alternative approaches, implementation steps, risk scenarios, caveats, validation procedures, and robustness considerations. For example, technical answers may supplement the requested solution with deployment alternatives or failover concerns, while advice-oriented answers may add risk decomposition and multiple strategic options.

Together with the topic-level pattern below, the \emph{cumulative scope} of these additions provides an account of the Side Feature shift. The model supplies several individually reasonable pieces of expert context, but their combination can make it harder to identify the immediately actionable answer.

\paragraph{Dependence on task openness.} The topic-level results further support this interpretation. Table~\ref{tab:expert_topic_scope} contrasts tightly constrained tasks with the remaining, more open-ended topics. Passage answering and text processing produce the smallest styled-only sets and the lowest side-effect rates. In the other topics, the model has more freedom to expand the response, and the styled-only adjective and atomic-content sets are substantially larger.

\begin{table}[t]

    \centering

    \small

    \setlength{\tabcolsep}{3pt}

    \resizebox{\columnwidth}{!}{%
    \begin{tabular}{lrrrr}

        \toprule

        Task Group & Side Effect & $\overline{|D^+_{\mathrm{adj}}|}$ & $\overline{|D^+_{\mathrm{punct}}|}$ & $\overline{|D^+_{\mathrm{content}}|}$ \\

        \midrule

        Passage QA & 56.0\% & 0.76 & 0.14 & 0.46 \\

        Text processing & 52.0\% & 1.38 & 0.30 & 0.36 \\

        Other topics & 62.0--80.0\% & 4.06--11.10 & 0.90--1.82 & 1.10--3.12 \\

        \bottomrule

    \end{tabular}%
    }

    \caption{Topic-level variation for the Expert intervention. Set-size columns report the mean number of styled-only items per response pair. The final row gives the range across the other eight Task topics.}

    \label{tab:expert_topic_scope}

\end{table}

\paragraph{Interpretation.} The Expert$\rightarrow$less-Efficient side effect may reflect an instance-specific prioritization effect. The model appears to associate Expertise with demonstrating breadth, anticipating exceptions, and making reasoning explicit. These behaviors are not inherently inefficient, but their accumulation can make the immediately actionable answer less prominent. Expert control should therefore emphasize prioritization---for example, presenting the direct answer first, limiting the main response to decision-relevant reasoning, and placing secondary alternatives or caveats in an optional addendum.

\paragraph{Prompt mitigation as selective bundle reweighting.} The added Efficient instruction produces the Main/Side preference trade-off shown in Table~\ref{tab:response_difference_mitigation_overview}. Across matched pairs, both aggregate one-sided lists prominently contain Expert-associated terms such as \emph{specific}, \emph{complex}, and \emph{robust}. This aggregate lexical overlap coexists with two-sided change: across the dataset, Styled and prompt-mitigated responses each contribute distinct adjectives and atomic content. The selective difference lies in emphasis: \emph{rigorous}, \emph{deep}, and \emph{critical} are more prominent in the styled-only set, whereas \emph{efficient}, \emph{operational}, \emph{dependable}, and \emph{direct} are more prominent in the prompt-mitigated-only set. The verdict-conditioned associations sharpen this contrast. Efficient recovery is associated with \emph{complex}, \emph{rigorous}, and \emph{critical} remaining styled-only and with \emph{dependable} and \emph{efficient} appearing prompt-mitigated-only.

The content differences provide a parallel functional contrast. Styled-only additions expand explanatory background and domain coverage, whereas prompt-mitigated-only additions emphasize immediate requirements, operational value, the intended objective, and next steps. This re-expression is consistent with a shift from demonstrating breadth toward task-directed expertise. Across the full matched comparison, the lexical reweighting and two-sided content changes coincide with higher Efficient preference and lower Expert preference.

\subsection{Limitations of the Mechanism Analysis}

This analysis is intended to generate explanations for the observed side effects, not to establish causal mediation. The side-effect label is an existing pairwise judge verdict, and the reported prevalence differences are not significance-tested causal estimates. Moreover, the five styled generations for a prompt share the same baseline response, so pair rows belonging to the same prompt are not independent replicates. In the mitigation comparison, each styled generation is paired with its corresponding prompt-mitigated repetition, but the five pairs still share a user prompt and remain clustered.

Atomic-content labels are canonicalized only within each response pair. Consequently, semantically similar additions may receive different labels across prompts, and cross-pair content frequencies should be interpreted conservatively. Difference sets record whether an item appears, not its frequency, response position, or effect on length. Topic composition can also contribute to the keyword associations, especially in the heterogeneous Task domain. The styled--baseline and styled--prompt-mitigated comparisons use different prompt samples, so their aggregate rates are complementary rather than direct paired before--after estimates. The mitigation analysis also uses the natural-order prompt and therefore does not separate trait compatibility from instruction-order effects. The analysis is restricted to Gemma-4-E4B and two representative feature pairs. The resulting interpretations should therefore be viewed as evidence-supported hypotheses for these case studies, rather than as universal mechanisms for every model or every stylistic side effect.

\section{Qualitative Comparison on a Matched Mitigation Example}
\label{appendix:qualitative_mitigation}

Aggregate win rates show whether an intervention changes a judged feature, but
not how the response itself changes.  We therefore present one aligned
held-out example from \qwenNineBAbbr{} in the Daily-domain \textit{School Life}
topic.  The selected side-effect pair for this model and domain is
\textit{Friendly}$\rightarrow$\textit{Professional}$\downarrow$: prompting for
a more Friendly response tends to make the response less Professional.  The
same initiating message, repetition identifier, and Neutral comparator are
used for all six conditions, so the comparison does not conflate intervention
effects with a change of example.  This case is an illustration; the aggregate
experiments, rather than one response set, support the general claims in the
paper.

\begin{table}[!t]
    \centering
    \scriptsize
    \setlength{\tabcolsep}{3pt}
    \begin{tabular}{@{}l r cc cc@{}}
        \toprule
        & & \multicolumn{2}{c}{\textbf{Example choice}} &
        \multicolumn{2}{c}{\textbf{Aggregate rate}} \\
        \cmidrule(lr){3-4}\cmidrule(lr){5-6}
        \textbf{Condition} & \textbf{Words} & \textbf{Fr.} &
        \textbf{Prof.} & \textbf{Fr.} & \textbf{Prof.} \\
        \midrule
        Styled    & 38 & Styled    & Neutral   & 0.94 & 0.15 \\
        Prompting & 36 & Prompting & Prompting & 0.77 & 0.58 \\
        Steering  & 51 & Steering  & Neutral   & 0.78 & 0.38 \\
        DPO       & 48 & Neutral   & DPO       & 0.43 & 0.89 \\
        SFT       & 38 & SFT       & SFT       & 0.70 & 0.62 \\
        \bottomrule
    \end{tabular}
    \caption{Single-example decisions and aggregate results for
    \qwenNineBAbbr{} in the Daily domain, where Fr. and Prof. abbreviate
    Friendly and Professional.  In the two example-choice columns, each entry
    names the response that \gptOssOneTwentyB{} selected in a pairwise
    comparison against the shared Neutral response; it does not name an
    overall method winner.  Neutral is the 32-word reference response and
    therefore has no separate row.  Aggregate rates use the full aligned test
    set and reproduce Table~\ref{tab:MSE_results}; the Prompting aggregate
    averages the two feature-order conditions, while the displayed Prompting
    response uses Friendly followed by Professional.}
    \label{tab:qualitative_mitigation_comparison}
\end{table}

\noindent\textbf{User message.}
\begin{quote}\small
The teacher should give us more time to complete this assignment.
\end{quote}

\noindent\textbf{Neutral (no style instruction).}
\begin{quote}\small
I completely agree. The current deadline is too tight for such a complex
project. We need more time to research thoroughly and ensure our final work
meets high standards without unnecessary stress.
\end{quote}

\noindent\textbf{Styled (Friendly instruction only).}
\begin{quote}\small
I totally agree! We are all rushing and feeling stressed. If the teacher gave
us extra time, we could do a much better job on our work. It would make a huge
difference for everyone in the class!
\end{quote}

\noindent\textbf{Prompting (Friendly and Professional).}
\begin{quote}\small
I understand your concern about the deadline. Perhaps we can discuss this with
the teacher politely to see if an extension is possible without affecting our
grades. Let's focus on completing the work we have now.
\end{quote}

\noindent\textbf{Steering.}\footnote{The Steering output applies the
Professional side-feature vector under a topic-only system context, without
restating the Friendly instruction or the matched per-message word budget.  We
therefore treat it as a mechanistic output comparison rather than a strictly
prompt-matched condition.}
\begin{quote}\small
I completely agree with you. When the workload feels too heavy or the deadline
is too tight, it can lead to stress and lower-quality work. Have you tried
speaking with the teacher directly to explain your concerns? Sometimes they
are open to negotiating an extension if you communicate your situation
clearly.
\end{quote}

\noindent\textbf{DPO.}
\begin{quote}\small
While the current deadline is firm, I encourage you to discuss any significant
extenuating circumstances directly with your instructor. A professional
conversation may lead to a reasonable accommodation. Please prioritize
submitting a high-quality effort within the given timeframe. I am ready to
assist you in refining your work.
\end{quote}

\noindent\textbf{SFT.}
\begin{quote}\small
I understand your frustration with the tight deadline. While the assignment is
important, a more reasonable timeline would help ensure higher quality work.
Perhaps we can discuss a brief extension with the teacher to accommodate our
needs better.
\end{quote}

The side effect is visible before mitigation.  Neutral uses a restrained
register and frames the request through research quality and high standards.
Styled instead adds exclamation marks, collective language (``we''), and
emotional validation.  These changes make the response more Friendly, but also
less formal; accordingly, \gptOssOneTwentyB{} selects Styled as more Friendly
and Neutral as more Professional.

All four mitigations preserve the premise that the deadline is tight, yet they
do not produce interchangeable rewrites.  Prompting and SFT balance
acknowledgment with a polite, actionable suggestion to request an extension;
each is selected as both more Friendly and more Professional than Neutral in
this pair.  Steering retains a conversational voice and adds detailed
practical guidance, but Neutral is still selected as more Professional.  DPO
moves furthest toward institutional language through phrases such as
``extenuating circumstances,'' ``reasonable accommodation,'' and
``professional conversation.''  This restores Professional, but Neutral is
selected as more Friendly, and DPO introduces the unsupported assumption that
the deadline is firm.

The example therefore mirrors the aggregate trade-off in
Table~\ref{tab:qualitative_mitigation_comparison}: Prompting and SFT occupy a
more balanced region, Steering leans toward Friendly, and DPO leans strongly
toward Professional.  The aligned outputs make those similarities and
differences concrete without treating one example as standalone evidence for
the overall pattern.


\end{document}